\documentclass[lettersize,journal]{IEEEtran}
\usepackage[colorlinks=true,linkcolor=red,citecolor=cvprblue,urlcolor=blue]{hyperref}
\usepackage{amsmath,amsfonts}
\usepackage{algorithmic}
\usepackage{algorithm}
\usepackage{array}
\usepackage[caption=false,font=normalsize,labelfont=sf,textfont=sf]{subfig}
\usepackage{textcomp}
\usepackage{stfloats}
\usepackage{url}
\usepackage{verbatim}
\usepackage{graphicx}
\usepackage{cite}
\usepackage{orcidlink}
\usepackage[algo2e,ruled,linesnumbered]{algorithm2e}
\SetAlgorithmName{Alg.}{Algorithm}{List of Algorithms}

\makeatletter
\renewcommand{\fnum@algocf}{%
  \normalfont\mdseries\footnotesize
  \algorithmcfname\nobreakspace\thealgocf
}
\renewcommand{\algocf@captiontext}[2]{%
  \algocf@captionlayout{%
    \parbox{\dimexpr\linewidth-\algomargin\relax}{%
      \normalfont\footnotesize
      #1.\quad #2\par
    }%
  }%
}
\makeatother
\SetAlCapSkip{0.5\baselineskip}

\usepackage{hhline}
\usepackage{booktabs}
\usepackage{colortbl}
\usepackage{pifont}
\usepackage{makecell}
\usepackage{multirow}
\usepackage{tabularx}
\usepackage[accsupp]{axessibility}
\usepackage[capitalise]{cleveref}
\crefname{section}{Sec.}{Secs.}
\Crefname{section}{Section}{Sections}
\crefname{table}{Tab.}{Tabs.}
\Crefname{table}{Table}{Tables}
\crefname{figure}{Fig.}{Figs.}
\Crefname{figure}{Figure}{Figures}
\crefname{algorithm}{Alg.}{Algs.}
\Crefname{algorithm}{Algorithm}{Algorithms}
\definecolor{cvprblue}{rgb}{0.21,0.49,0.74}
\definecolor{MyGray}{rgb}{0.85, 0.85, 0.85}

\newcommand{\xt}[1]{\textit{#1}}

\newcommand{\etal}{\xt{et al.\ }}
\definecolor{Gray}{rgb}{0.35, 0.35, 0.35}
\definecolor{DarkGray}{rgb}{0.8, 0.8, 0.8}

\begin{document}

\newcommand{\makeorcid}[1]{\hspace{-1mm}{~\orcidlink{#1}}}

\newcommand{\wgc}[1]{\textcolor{blue}{[wgc: #1]}}
\newcommand{\czx}[1]{\textcolor{red}{[czx: #1]}}

\title{EvTrajGS: Accurate and Efficient 3D Gaussian Splatting from Unposed Event Streams}

\author{
Zixuan~Chen,~
Jiakai~Zhang,~
Junhao~Dong,~
Guangcong~Wang,~
Jianhuang~Lai,~\IEEEmembership{Senior~Member,~IEEE},~
Yew-Soon~Ong,~\IEEEmembership{Fellow,~IEEE},~
and~Xiaohua~Xie
\thanks{Manuscript received XXX XX, XXXX; revised XXX XX, XXXX. This work was supported in part by National Natural Science Foundation of China (U22A2095) and the Project of Guangdong Provincial Key Laboratory of Information Security Technology (2023B1212060026).}
\thanks{(Corresponding author: Xiaohua Xie.)}
\thanks{Zixuan Chen, Jiakai Zhang, Jianhuang Lai and Xiaohua Xie are with the School of Computer Science and Engineering, Sun Yat-sen University, Guangzhou 510006, Guangdong, China; and with the Guangdong Province Key Laboratory of Information Security Technology, Guangzhou 510006, Guangdong, China; and also with the Key Laboratory of Machine Intelligence and Advanced Computing, Ministry of Education, Guangzhou 510006, Guangdong, China. (e-mail: \{chenzx3, zhangjk8\}@mail2.sysu.edu.cn; \{stsljh, xiexiaoh6\}@mail.sysu.edu.cn)}
\thanks{Junhao Dong and Yew-Soon Ong are with the College of Computing and Data Science, Nanyang Technological University, Singapore, Singapore. (e-mail: \{junhao003, asysong\}@ntu.edu.sg)}
\thanks{Guangcong Wang is with the School of Computing and Information Technology, Great Bay University, Dongguan 523000, Guangdong, China. (e-mail: wanggc3@gmail.com)}
}

\markboth{IEEE TRANSACTIONS ON PATTERN ANALYSIS AND MACHINE INTELLIGENCE, VOL. XX, NO. XX, XXX. XXXX}%
{Shell \MakeLowercase{\textit{Chen et al.}}: EvTrajGS: accurate and efficient 3D Gaussian Splatting from Unposed Event Streams}

\maketitle

\begin{abstract}
Event cameras, with high temporal resolution, high dynamic range, and asynchronous sensing characteristics, have shown great potential for dense 3D reconstruction. Traditional reconstruction methods based on off-the-shelf pose estimates achieve high efficiency but produce low-fidelity results, as inaccurate pose initialization introduces cumulative reconstruction errors. In contrast, recent SLAM-style methods stabilize joint pose-scene optimization through incremental tracking and mapping, yielding higher reconstruction fidelity at the expense of considerable computational overhead. To address this trade-off, this paper presents \textit{EvTrajGS}, an accurate and efficient 3D Gaussian Splatting framework for unposed event streams. Our method enables reliable joint pose-scene optimization initialized from coarse pose priors, eliminating the need for computationally expensive SLAM-style pipelines. \textit{EvTrajGS} parameterizes camera motion as a continuous-time trajectory initialized from discrete camera poses, providing a unified representation for pose refinement. We then aggregate adjacent trajectory states into a temporally coupled pose, promoting temporally consistent pose updates during joint optimization. Additionally, we introduce a loss-reweighted event sampling strategy to adaptively emphasize temporally under-reconstructed intervals. Extensive experiments on both synthetic and real-world datasets demonstrate that \textit{EvTrajGS} outperforms state-of-the-art methods in terms of both geometric reconstruction quality and pose estimation accuracy, achieving 3.8 dB higher PSNR, 0.1 higher SSIM, and over 40\% lower ATE RMSE while retaining high computational efficiency.
\end{abstract}


\section{Introduction}\label{sec:intro}
\IEEEPARstart{E}{vent} cameras are bio-inspired sensors characterized by low power consumption, high dynamic range, and asynchronous operation, capturing pixel-wise brightness changes with microsecond-level temporal resolution ($10^{-6}$ s).
Compared with conventional frame-based cameras, they provide significantly higher temporal resolution with negligible motion blur, stronger robustness under challenging illumination, and substantially lower power consumption with reduced data redundancy.
These unique sensing characteristics provide rich temporal motion cues and efficient visual observations, offering new opportunities for dense 3D reconstruction.

\begin{figure}[!t]
  \includegraphics[width=\linewidth]{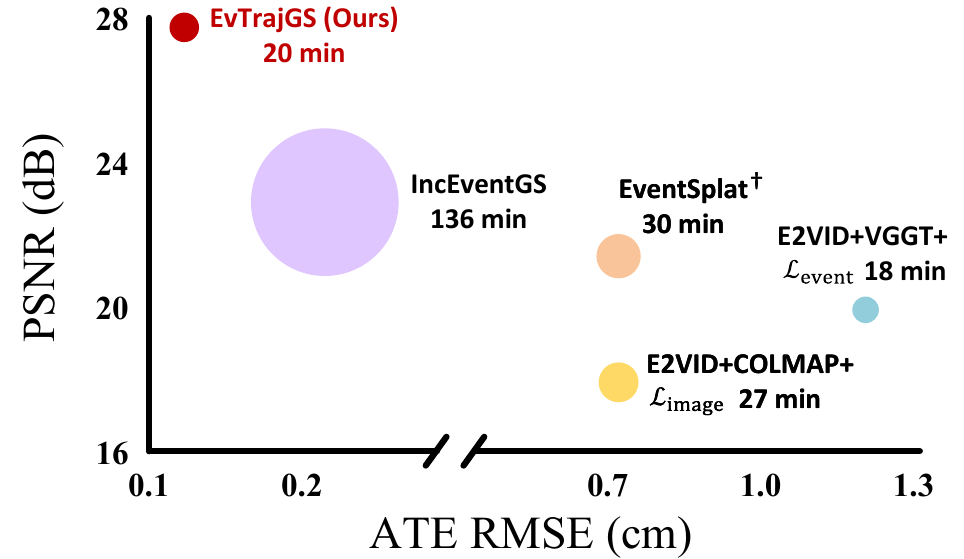}
  \caption{
    Performance of state-of-the-art methods on the Replica dataset \cite{2019replica}, including the fixed-pose reconstruction method EventSplat$^\dag$ \cite{yura2025eventsplat}, and the SLAM-style joint optimization method IncEventGS \cite{2025inceventgs}.
    The radius of circles is proportional to their runtime evaluated on a single NVIDIA RTX 3090 GPU.
    EventSplat$^\dag$ is reproduced following its described pipeline based on E2VID \cite{2019e2vid}, COLMAP \cite{2016colmap}, and $\mathcal{L}_{\text{event}}$, as its code is not publicly available.
    Starting from E2VID \cite{2019e2vid} + VGGT \cite{wang2025vggt}, \textit{EvTrajGS} achieves the best reconstruction quality and pose accuracy among all compared methods, while being nearly 7$\times$ faster than the SLAM-style method IncEventGS.
  }
  \label{fig:time}
\end{figure}

Recent advances in neural scene representations have unlocked the potential of event cameras for dense 3D reconstruction. 
Neural Radiance Fields (NeRF) \cite{2022nerf} and 3D Gaussian Splatting (3DGS) \cite{3dgs} make it possible to reconstruct dense 3D scenes from the sparse and asynchronous measurements of event cameras. 
Pioneering studies \cite{2023e_nerf,2023ev-nerf,2023eventnerf} demonstrated high-fidelity NeRF reconstruction under an event-based objective, while subsequent work \cite{han2024event3dgs} extended this paradigm to 3DGS, achieving faster convergence and better visual quality.

\begin{figure*}[!t]
  \includegraphics[width=\textwidth]{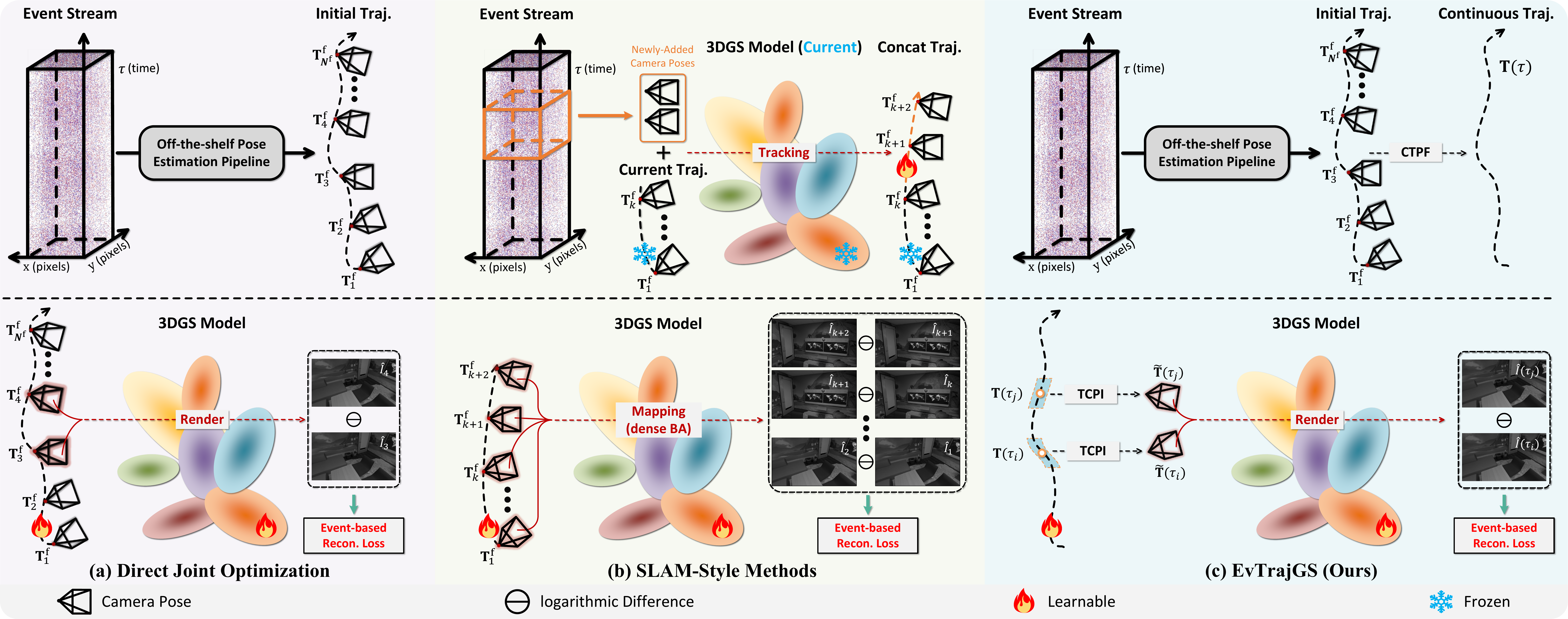}
  \caption{
  Comparison of three representative paradigms, which differ in how camera poses are estimated, represented, and refined.
  \textbf{(a)} Direct joint optimization treats off-the-shelf pose estimates as discrete learnable variables and jointly optimizes them with the scene representation.
  Without effective mechanisms for coordinating poses, this paradigm may produce unreliable pose updates due to pose-scene ambiguity under the event-based supervision, leading to suboptimal results.
  \textbf{(b)} SLAM-style methods stabilize joint optimization through an incremental tracking-and-mapping pipeline, where tracking estimates newly added camera poses and mapping jointly refines a local window of camera poses and the scene representation via dense BA.
  Although effective, this paradigm incurs substantial computational overhead from repeated tracking-and-mapping and dense BA.
  \textbf{(c)} \textit{EvTrajGS} first parameterizes the off-the-shelf pose estimates as a continuous-time trajectory function using CTPF, and then aggregates neighboring trajectory states into temporally coupled pose through TCPI.
  This captures local motion trends beyond isolated pose variables, enabling efficient and reliable joint optimization over the entire sequence.
  }
  \label{fig:motivation}
\end{figure*}

Despite this progress, most existing methods assume camera trajectories are available during reconstruction. 
Such trajectories are typically obtained under carefully controlled acquisition conditions and are often unavailable in practical scenarios, where event streams are captured without accurate camera poses. 
The reliance on known trajectories limits the applicability of existing methods to practical scenarios. 
This raises a key question: \textit{How can we achieve accurate and efficient dense 3D reconstruction from unposed event streams?}

As shown in \cref{fig:time}, recent methods face a trade-off between reconstruction accuracy and computational efficiency.
To further understand, we analyze three representative paradigms:

\noindent\textbf{First}, fixed-pose reconstruction methods estimate camera trajectories using off-the-shelf pipelines and keep them fixed during scene reconstruction.
For example, EventSplat \cite{yura2025eventsplat} combines E2VID \cite{2019e2vid} and COLMAP \cite{2016colmap} to obtain camera poses for 3DGS reconstruction.
Although it is computationally efficient, its performance is limited by the accuracy of the estimated camera poses.
The estimated trajectories may contain local inaccuracies and fluctuations due to artifacts in the reconstructed intensity videos.
Since these poses remain fixed during reconstruction, such errors cannot be corrected and may cause the scene representation to compensate for inaccurate viewpoints, leading to suboptimal reconstructions.

\noindent\textbf{Second}, as shown in \cref{fig:motivation} \textbf{(a)}, direct joint optimization treats camera poses as isolated learnable variables and jointly optimizes them with the scene representation.
Although it allows initial pose errors to be corrected, event-based supervision primarily constrains the accumulated log-intensity changes between timestamps.
Consequently, pose and scene parameters can compensate for each other under weak event constraints, making these pose variables prone to inconsistent updates and limiting further pose refinement and reconstruction quality.
A common solution to alleviate this issue is to introduce first- or second-order pose difference constraints to enforce local trajectory smoothness.
However, this may conflict with the reconstruction objective, resulting in suboptimal optimization.

\noindent\textbf{Third}, recent methods perform joint pose-scene optimization through SLAM-style pipelines (\textit{e.g.,} IncEventGS \cite{2025inceventgs}, E2EGS \cite{kim2026e2egs}, as shown in \cref{fig:motivation} \textbf{(b)}).
These methods alternate between tracking and mapping in an incremental manner, and perform dense bundle adjustment (BA) over multiple event intervals, enforcing cross-view geometric consistency to coordinate the pose updates during the mapping stage.
Therefore, camera trajectories and scene representations are progressively improved, enabling accurate and high-fidelity reconstruction.
However, the incremental tracking-and-mapping process and dense BA over multiple event intervals require repeated rendering and joint optimization, resulting in substantial computational cost.

In this paper, we propose \textit{EvTrajGS}, an accurate and efficient framework for 3D Gaussian Splatting from unposed event streams.
As shown in \cref{fig:motivation} \textbf{(c)}, \textit{EvTrajGS} enables reliable joint optimization starting from off-the-shelf pose estimates without relying on costly SLAM-style pipelines.
Motivated by the continuity of camera motion, neighboring poses within a short temporal window typically exhibit correlated motion patterns.
Our key idea is to exploit this property by coupling neighboring trajectory states rather than optimizing isolated pose variables, thereby capturing local motion trends and promoting temporally consistent pose updates.

To achieve this goal, we first introduce a \textbf{Continuous Time-to-Pose Function} (CTPF), which parameterizes camera motion as a continuous-time trajectory initialized from discrete pose estimates provided by an off-the-shelf pipeline, enabling pose queries at arbitrary timestamps within a local temporal window.
Under the small-motion assumption, the neighboring relative motions can then be integrated in the Lie algebra space using the first-order Baker-Campbell-Hausdorff (BCH) approximation.
Building upon this formulation, we propose \textbf{Temporally Coupled Pose Integration} (TCPI), which employs a time-adaptive truncated Gaussian weighting function to aggregate neighboring trajectory states within a local temporal window into a temporally coupled pose for rendering.
Compared with dense BA, which jointly optimizes over multiple event intervals in each iteration, TCPI optimizes only a single event interval per iteration, significantly reducing computational overhead.
Next, we jointly optimize the CTPF and 3DGS parameters under an event-based reconstruction objective. 
Furthermore, uniform event sampling may repeatedly select well-fitted intervals while under-sampling more difficult ones.
We therefore introduce a loss-reweighted event sampling strategy that adaptively emphasizes underfitted event intervals for more effective supervision during optimization.

Experiments on synthetic and real-world datasets demonstrate that \textit{EvTrajGS} outperforms state-of-the-art methods in both reconstruction quality and pose accuracy, achieving 3.8 dB higher PSNR, 0.1 higher SSIM, and over 40\% lower ATE RMSE.
Moreover, as shown in \cref{fig:time}, \textit{EvTrajGS} delivers high-fidelity reconstructions within 20 minutes on a single NVIDIA RTX 3090 GPU, achieving nearly a 7$\times$ speedup over the recent SLAM-style method IncEventGS \cite{2025inceventgs}, demonstrating its ability to break the accuracy-efficiency trade-off.

Overall, the main contributions are summarized as follows:

\begin{itemize}

\item We propose \textit{EvTrajGS}, an accurate and efficient framework for 3DGS reconstruction from unposed event streams, enabling reliable joint pose-scene optimization without costly incremental tracking and dense BA.

\item We parameterize isolated pose estimates as a continuous-time trajectory, and further adaptively aggregate neighboring trajectory states into a temporally coupled pose.
Together with loss-reweighted event sampling that emphasizes underfitted event intervals, these designs enable reliable and effective joint optimization.

\item Experiments on synthetic and real-world datasets demonstrate that \textit{EvTrajGS} outperforms state-of-the-art methods in both reconstruction quality and pose accuracy, achieving 3.8 dB higher PSNR, 0.1 higher SSIM, and over 40\% lower ATE RMSE while maintaining high efficiency.

\end{itemize}

\section{Related Work}\label{sec:rw}
\noindent\textbf{RGB-based 3D Representation.}
Neural radiance field (NeRF) \cite{2022nerf} models scenes implicitly using multi-layer perceptrons (MLPs) combined with volumetric rendering \cite{rendering} and alpha compositing \cite{alpha}.
NeRF has been extended to sparse-view reconstruction \cite{2023Sparsenerf}, fast inference \cite{2022instngp}, generative modeling \cite{2021Giraffe}, text-to-3D synthesis \cite{2023ProlificDreamer}, medical image reconstruction \cite{cunerf}, and anti-aliasing \cite{2022mip-nerf-360}.
In contrast, 3D Gaussian Splatting (3DGS) \cite{3dgs} explicitly represents scenes as a collection of 3D Gaussians and yields views via splatting \cite{splatting}, offering faster optimization and rendering.
3DGS has been applied to text-to-3D generation \cite{zhu2025segmentdreamer}, avatar synthesis \cite{2024splattingavatar}, single-view reconstruction \cite{szymanowicz2024splatter}, watermarking \cite{guardsplat} and SLAM \cite{matsuki2024gaussian}.

\noindent\textbf{Event-based 3D Reconstruction.}
Previous works generally assume known camera trajectories during optimization. 
Under this assumption, early approaches \cite{2023e_nerf,2023ev-nerf,2023eventnerf} first extended NeRF-based reconstruction to event streams. 
Low \etal \cite{2023robust_e_nerf} further improve robustness under sparse and noisy events, while Li \etal \cite{li2024benerf} combine event data with single blurry images for improved supervision.
More recently, 3D Gaussian representations have been introduced for event-based reconstruction \cite{wang2024evggs,han2024event3dgs,yura2025eventsplat,deguchi2024e2gs,lee2025diet-gs}. 
Some methods \cite{wang2024evggs,han2024event3dgs} rely on event streams for static scene reconstruction, whereas others \cite{deguchi2024e2gs,lee2025diet-gs} additionally incorporate blurry images for stronger supervision.
For unposed event streams, EventSplat \cite{yura2025eventsplat} performs fixed-pose reconstruction by combining E2VID \cite{2019e2vid} and COLMAP \cite{2016colmap} for pose estimation. 
However, such fixed-pose pipelines are highly sensitive to pose errors, leading to degraded reconstruction quality.
To address this limitation, recent methods \cite{2025inceventgs,kim2026e2egs} adopt a SLAM-style incremental tracking-and-mapping paradigm for joint optimization of camera poses and scene representations. 
For each incoming event segment, tracking estimates newly added camera poses against the current scene representation while keeping the scene parameters fixed, providing more reliable pose estimates for subsequent mapping.
Mapping then jointly refines the newly estimated poses, historical poses, and scene representation through dense bundle adjustment (BA).
SLAM-style pipelines produce more reliable pose updates and improve reconstruction fidelity, but require repeated rendering and iterative local optimization, resulting in substantial computational cost.

Unlike SLAM-style approaches \cite{2025inceventgs,kim2026e2egs}, \textit{EvTrajGS} adopts a different optimization paradigm by representing camera motion estimated by off-the-shelf pipelines as a continuous trajectory and enforcing temporal consistency during joint pose-scene optimization. 
This formulation enables stable optimization without relying on costly incremental tracking and dense BA, achieving superior reconstruction quality and pose accuracy with significantly reduced runtime.

\section{Preliminaries}\label{sec:preliminary}
\noindent\textbf{Event Generation Model.}
Event cameras are bio-inspired sensors that can asynchronously capture the ``events'' (\textit{i.e.,} brightness changes) of each independent pixel, which are widely used for numerous vision tasks \cite{ev-survey,ev_depth_survey,ev_slam_survey}.
Unlike conventional frame cameras that collect images at a fixed frame rate, event cameras can record events with microsecond-level temporal resolution ($10^{-6}$ s) to produce the event stream:
\begin{equation}
  \mathcal{E}=\{e_i\}^{N^\text{e}}_{i=1},\ \text{where}\ e_i=\{x_i,y_i,\tau_i,p_i\},\ \text{and}\ i\in\mathbb{N}^+,
\end{equation}
where $(x_i, y_i)$ denotes the pixel location, $\tau_i$ is the timestamp, and $p_i\in\{-1,+1\}$ is the polarity, indicating negative and positive brightness changes, respectively.
Specifically, whenever a change in the logarithmic intensity $\log(\mathbf{I})$ surpasses the threshold $C$, the event is triggered as:
\begin{equation}
  p=
  \begin{cases}
    & \!\!\!\!\!+1,\ \log(\mathbf{I}_{x,y}(\tau+\Delta\tau))-\log(\mathbf{I}_{x,y}(\tau)) \geq C \\
    & \!\!\!\!\!-1,\ \log(\mathbf{I}_{x,y}(\tau+\Delta\tau))-\log(\mathbf{I}_{x,y}(\tau)) \leq -C, \\
    & 0,\ \text{no\ events}
  \end{cases}
\end{equation}
where $\mathbf{I}$ denotes the intensity images and $\Delta\tau$ is a time interval between two adjacent timestamps of position $(x,y)$.

\noindent\textbf{3D Gaussian Splatting.}
3DGS \cite{3dgs} represents a scene using a set of anisotropic 3D Gaussians, each parameterized by mean $\boldsymbol{\mu}\in\mathbb{R}^3$, covariance $\boldsymbol{\Sigma}\in\mathbb{R}^{7}$, color $\boldsymbol{c} \in \mathbb{R}^{3}$ transformed from $k$-order spherical harmonics $\boldsymbol{h} \in \mathbb{R}^{3\times (k+1)^2}$, and opacity $\alpha \in \mathbb{R}^1$.
A Gaussian at position $\mathbf{x}$ is defined as:
\begin{equation}
\label{eq:gaussian}
\mathcal{G}\left(\mathbf{x}:\boldsymbol{\mu},\boldsymbol{\Sigma}\right)=\exp\left(-\frac{1}{2}(\mathbf{x}-\boldsymbol{\mu})^{\top}\boldsymbol{\Sigma}^{-1}(\mathbf{x}-\boldsymbol{\mu})\right).
\end{equation}
Given the camera pose $\mathbf{T}\in \text{SE}(3)$ and the intrinsic matrix $\mathbf{K}\in\mathbb{R}^{3\times3}$, the projection matrix $\mathbf{P}=\mathbf{K}[\mathbf{I}_3|0]$ and the camera transformation $\mathbf{W}=\mathbf{T}^{-1}$ are first derived.
Let $\mathbf{J}$ denote the Jacobian of the affine approximation to the projection, the 2D projection of a 3D Gaussian with mean $\boldsymbol{\mu}$ and covariance $\boldsymbol{\Sigma}$ is then computed as $\hat{\boldsymbol{\mu}}=\mathbf{P}\mathbf{W}\boldsymbol{\mu}$ and $\hat{\boldsymbol{\Sigma}}=\mathbf{J}\mathbf{W}\boldsymbol{\Sigma}\mathbf{W}^{\top}\mathbf{J}^{\top}$.
Pixels are then rendered via alpha compositing:
\begin{equation}
\hat{\mathbf{I}}_{x,y}\! = \!\!\sum_{i=1}^{N^\text{g}}\!\boldsymbol{c}_{i}\eta_{i}\!\prod_{j=1}^{i-1}(1-\eta_{j}),\ \text{where}\ 
\eta_i\!=\!\alpha_i\mathcal{G}((x,y)\!:\!\hat{\boldsymbol{\mu}}_i,\hat{\boldsymbol{\Sigma}}_i).
\label{eq:3dgs_rendering}
\end{equation}
The model can be optimized by an RGB reconstruction loss:
\begin{equation}
\mathcal{L}_\text{image}=\lambda_\text{ssim}\mathcal{L}_\text{ssim}(\hat{\mathbf{I}}, \mathbf{I})+(1-\lambda_\text{ssim})\mathcal{L}_1(\hat{\mathbf{I}}, \mathbf{I}),
\label{eq:rgb_loss}
\end{equation}
where $\mathbf{I}$ is the ground-truth image and $\lambda_\text{ssim}=0.2$.

\section{Method}
\begin{figure*}
  \includegraphics[width=\textwidth]{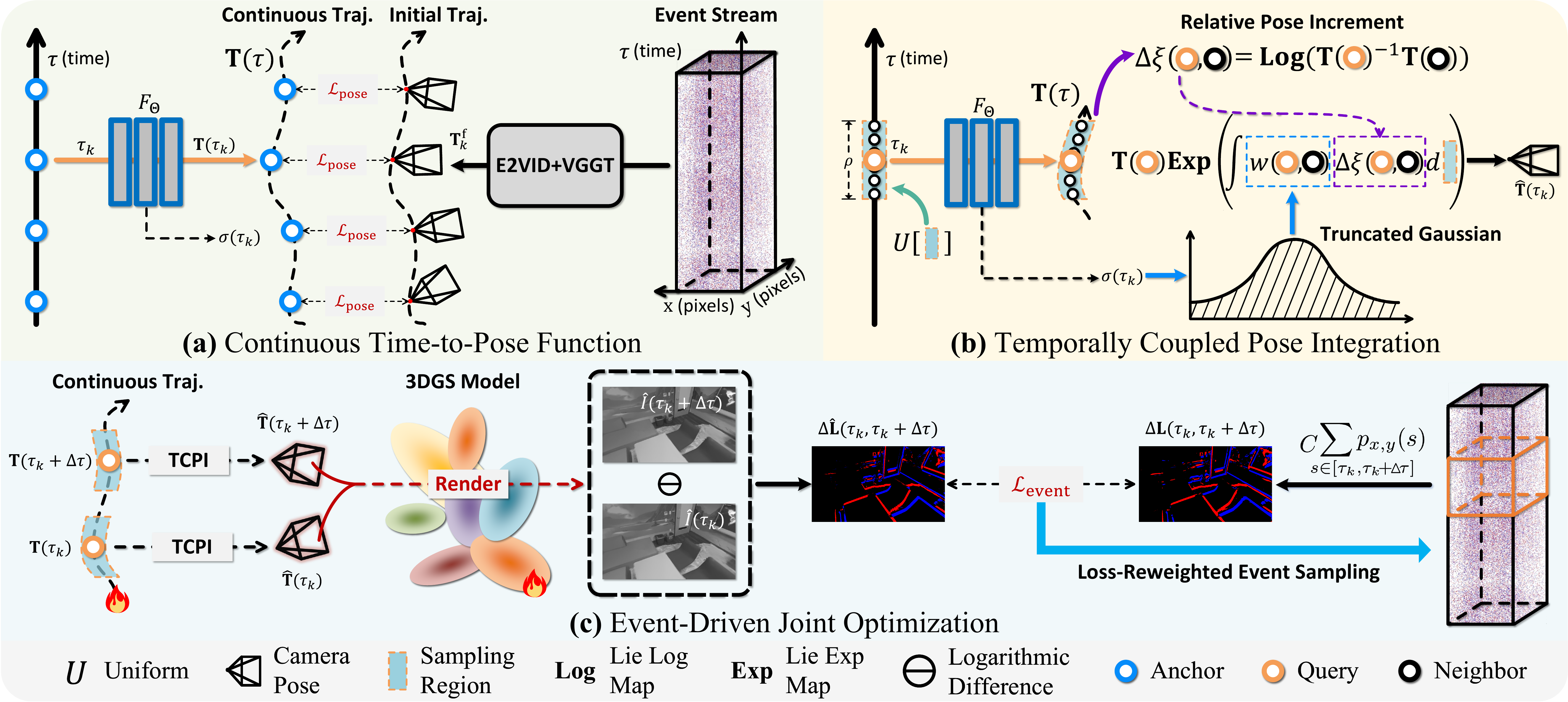}
  \caption{
  The overall framework of EvTrajGS.
  \textbf{(a)} Given an event stream, we represent camera motion as a continuous trajectory using an MLP $F_\Theta$, which maps timestamps $\tau$ to camera poses $\mathbf{T}(\tau)$.
  The continuous trajectory is initialized by minimizing $\mathcal{L}_{\text{pose}}$ in \cref{eq:loss_pose} between the predicted poses $\{\mathbf{T}(\tau_i^\text{f})\}_{i=1}^{N^\text{f}}$ and the corresponding initial poses $\{\mathbf{T}^{\text{f}}_i\}_{i=1}^{N^\text{f}}$, estimated by E2VID \cite{2019e2vid} and VGGT \cite{wang2025vggt}.
  \textbf{(b)} Given a query timestamp $\tau_k$, we uniformly sample the interval into $M$ temporal offsets $\{\delta_i\}_{i=1}^{M}$ within the $\rho$-length temporal window $[-\frac{\rho}{2}, \frac{\rho}{2}]$.
  By feeding $\tau_k$ and the sampled timestamps $\{\tau_k+\delta_i\}^M_{i=1}$ into $F_\Theta$, we obtain the relative pose increments $\{\Delta\boldsymbol{\xi}(\tau_k,\tau_k+\delta_i)\}_{i=1}^{M}$ in $\mathfrak{se}(3)$.
  The corresponding continuous weighting values $\{w(\tau_k,\tau_k+\delta_i)\}_{i=1}^{M}$ are computed according to the truncated Gaussian distribution in \cref{eq:truncated_gaussian} and normalized using \cref{eq:discrete_weight} to obtain the discrete integration weights $\{\hat{w}_{k,i}\}_{i=1}^{M}$.
  Finally, we numerically approximate the pose integral in \cref{eq:integral} using these discrete weights according to \cref{eq:numerical}, yielding the temporally coupled pose $\hat{\mathbf{T}}(\tau_k)$ for subsequent optimization.
  \textbf{(c)} Two nearby timestamps $\tau_k$ and $\tau_k'=\tau_k+\Delta\tau$ are randomly sampled to obtain the corresponding event measurement $\Delta\mathbf{L}(\tau_k,\tau_k')$ in \cref{eq:event_sum}.
  The intensity images $\hat{\mathbf{I}}(\tau_k)$ and $\hat{\mathbf{I}}(\tau_k')$ are rendered using the temporally coupled poses $\hat{\mathbf{T}}(\tau_k)$ and $\hat{\mathbf{T}}(\tau_k')$ through the 3DGS pipeline.
  We then compute the logarithmic difference $\Delta\hat{\mathbf{L}}(\tau_k,\tau_k')$ between the two rendered views, which is used to jointly optimize the CTPF and 3DGS parameters by minimizing the event-driven reconstruction loss in \cref{eq:event_gray}, together with the loss-reweighted event sampling module for effective supervision.
  }
  \label{fig:overall}
\end{figure*}

In this section, we propose \textit{EvTrajGS}, an accurate and efficient framework for 3DGS reconstruction from unposed event streams.
Starting from coarse camera poses estimated by off-the-shelf pipelines, \textit{EvTrajGS} jointly optimizes the camera trajectory and scene representation without relying on costly incremental tracking-and-mapping or dense bundle adjustment.
Specifically, we first introduce a Continuous Time-to-Pose Function (CTPF) to parameterize the discrete pose estimates as a continuous-time trajectory function.
This representation avoids independently optimizing isolated pose variables and enables pose queries at arbitrary timestamps.
Building upon CTPF, we present Temporally Coupled Pose Integration (TCPI), which adaptively aggregates neighboring trajectory states within a local temporal window into a temporally coupled pose for rendering.
By capturing the local motion trend beyond an isolated pose estimate, TCPI guides neighboring pose updates in a temporally consistent manner.
We then jointly optimize the CTPF and 3DGS parameters under event-based supervision.
To provide more effective supervision, we introduce a loss-reweighted event sampling (LRES) strategy that adaptively emphasizes underfitted event intervals.
Together, these components enable reliable and efficient joint optimization from off-the-shelf estimates, addressing the accuracy-efficiency trade-off.
The main notations are summarized in \Cref{tab:notations}, and the overview is depicted in \cref{fig:overall}.

\begin{table}[t]
\centering
\caption{Summary of the main notations used in this paper.}
\label{tab:notations}
\scriptsize
\begin{tabularx}{\columnwidth}{
    >{\centering\arraybackslash}p{0.12\columnwidth}
    X
}
\toprule
\textbf{Notation} & \textbf{Description} \\
\midrule

$\mathbf I(\tau),\,\hat{\mathbf I}(\tau)$
&
Observed and rendered intensity images at timestamp $\tau$, respectively.
\\

$\Delta\mathbf L,\,\Delta\hat{\mathbf L}$
&
Event-derived and rendered logarithmic difference maps, respectively.
\\

$F_\Theta$
&
CTPF parameterized by an MLP with parameters $\Theta$.
\\

$\sigma(\tau)$$\in$$\mathbb{R}$
&
Time-dependent temporal scale obtained from CTPF.
\\

$\mathbf q(\tau)$$\in$$\mathbb{R}^4$
&
Rotation (unit quaternion) predicted by CTPF at timestamp $\tau$.\\

$\mathbf t(\tau)$$\in$$\mathbb{R}^3$
&
Translation predicted by CTPF at timestamp $\tau$.
\\

$\mathbf T(\tau)$
&
Camera pose in $\text{SE}(3)$ predicted by CTPF at timestamp $\tau$.
\\

$\tau_i^\text{f},\,\mathbf T_i^\text{f}$
&
Timestamp and initial pose corresponding to the $i$-th intensity frame.
\\

$\rho$
&
Length of the local temporal integration window centered at $\tau_k$.
\\

$M$
&
Number of timestamps used for numerical pose integration.
\\

$\Delta\boldsymbol{\xi}(\tau_k,\tau)$
&
Relative pose increment in $\mathfrak{se}(3)$ between $\mathbf T(\tau_k)$ and $\mathbf T(\tau)$.
\\

$w(\tau_k,\tau)$
&
Continuous truncated-Gaussian integration density.
\\

$\hat w_{k,i}$
&
Normalized discrete integration weight assigned to $\delta_i$.
\\

$\tilde{\mathbf T}(\tau_k)$
&
Temporally coupled pose defined by continuous pose integration.
\\

$\hat{\mathbf T}(\tau_k)$
&
Numerical approximation of $\tilde{\mathbf T}(\tau_k)$ used for rendering.
\\
\bottomrule
\end{tabularx}
\end{table}

\subsection{Continuous Time-to-Pose Function}
\label{sec:ctpf}
\textit{EvTrajGS} starts from a set of discrete camera poses estimated by an off-the-shelf pipeline.
Directly treating these poses as independent variables neither provides a unified representation of camera motion nor supports pose queries at arbitrary timestamps required for local trajectory integration.
To address this limitation, we introduce a \textbf{Continuous Time-to-Pose Function} (CTPF), which parameterizes the camera motion as a shared continuous trajectory function from discrete pose estimates.
We next describe its continuous trajectory representation and initialization from the coarse pose estimates.

\noindent\textbf{Continuous Trajectory Representation.}
Given an arbitrary normalized timestamp $\tau\in[0,1]$, CTPF applies the Fourier positional encoding $\gamma(\tau)$ and feeds it into a multi-layer perceptron (MLP) $F_\Theta$ to predict an unconstrained quaternion vector, a translation vector, and a temporal-scale logit:
\begin{equation}
\bar{\mathbf{q}}(\tau),\mathbf{t}(\tau),s_{\sigma}(\tau)
=
F_\Theta\bigl(\gamma(\tau)\bigr).
\label{eq:mlp}
\end{equation}
The camera rotation is represented by the unit quaternion
\begin{equation}
\mathbf{q}(\tau)
=
\frac{\bar{\mathbf{q}}(\tau)}
{\|\bar{\mathbf{q}}(\tau)\|_2},
\label{eq:quaternion_normalization}
\end{equation}
while $\mathbf{t}(\tau)\in\mathbb{R}^3$ is directly used as the translation vector.
Together, $\mathbf{q}(\tau)$ and $\mathbf{t}(\tau)$ define the camera pose $\mathbf{T}(\tau)\in\text{SE}(3)$.
To obtain a positive and bounded temporal scale, we map the unconstrained output $s_{\sigma}(\tau)$ to a predefined range $[\sigma_{\min},\sigma_{\max}]$:
\begin{equation}
\sigma(\tau)=\sigma_{\min}+\bigl(\sigma_{\max}-\sigma_{\min}\bigr)\operatorname{Sigmoid}\!\left(s_{\sigma}(\tau)\right).
\label{eq:bounded_sigma}
\end{equation}
We zero-initialize the temporal-scale output head, such that $s_{\sigma}(\tau)=0$ for all timestamps at initialization, yielding $\sigma(\tau)=(\sigma_{\min}+\sigma_{\max})/2$.
The resulting time-dependent scale $\sigma(\tau)$ is subsequently used by TCPI to control the extent of local trajectory integration. 
Following \cite{2022nerf}, the Fourier positional encoding is defined as:
\begin{equation}
\gamma(\tau)
=
[\tau]
\cup
\bigcup_{i=0}^{\omega}
\left[
\sin(2^i\tau),
\cos(2^i\tau)
\right],
\label{eq:PE}
\end{equation}
where $\omega$ denotes the maximum frequency exponent and the encoding contains $\omega+1$ frequency bands.

\noindent\textbf{Continuous Trajectory Initialization.}
To initialize CTPF and the 3DGS representation, we partition the event stream into $N^\text{f}$ non-overlapping temporal segments with corresponding timestamps $\{\tau_i^\text{f}\}_{i=1}^{N^\text{f}}$.
Each segment is converted into an intensity frame using E2VID \cite{2019e2vid}, producing an intensity-frame sequence.
We then feed this sequence into VGGT \cite{wang2025vggt} to estimate an initial set of camera poses $\{\mathbf{T}_i^\text{f}\}_{i=1}^{N^\text{f}}$ and 3D points.
The estimated poses serve as trajectory anchors for initializing CTPF, while the 3D points initialize the 3DGS representations. Specifically, CTPF is initialized by minimizing
\begin{equation}
\mathcal{L}_\text{pose}
\!=\!
\frac{1}{N^\text{f}}
\!\sum_{i=1}^{N^\text{f}}
\left[
d_{\text{SO}(3)}
\!\left(
\mathcal{T}_{\mathbf{q}\rightarrow\mathbf{R}}
\bigl(\mathbf{q}(\tau_i^\text{f})\bigr),
\mathbf{R}_i^\text{f}
\right)
\!+\!
\left\|
\mathbf{t}(\tau_i^\text{f})
-
\mathbf{t}_i^\text{f}
\right\|_2
\right],
\label{eq:loss_pose}
\end{equation}
where $\mathbf{R}_i^\text{f}$ and $\mathbf{t}_i^\text{f}$ denote the rotation and translation components of $\mathbf{T}_i^\text{f}$, respectively.
$\mathcal{T}_{\mathbf{q}\rightarrow\mathbf{R}}$ converts a unit quaternion into its corresponding rotation matrix in $\text{SO}(3)$, and $d_{\text{SO}(3)}$ denotes the geodesic distance on the rotation manifold.

After initialization, CTPF is jointly optimized with the 3DGS representation under the event-based reconstruction objective.
Since camera poses at all timestamps are generated by the shared parameters $\Theta$, pose refinement is performed over a unified trajectory rather than a collection of independently optimized pose variables.
Moreover, CTPF provides the continuous pose and temporal-scale queries required to construct local trajectory segments at arbitrary timestamps.

\subsection{Temporally Coupled Pose Integration}
\label{sec:tcpi}
CTPF parameterizes camera motion as a continuous-time trajectory function initialized from discrete pose estimates and enables pose queries at arbitrary timestamps.
However, directly using an isolated pose for rendering remains susceptible to local trajectory fluctuations.
Motivated by the continuity of camera motion, neighboring poses within a short temporal window typically follow a correlated local motion trend.
Integrating these neighboring poses into a single pose can therefore better capture the local motion trend and mitigate the influence of isolated pose inaccuracies and fluctuations.
Since the corresponding relative motions are generally small within a local temporal window, they can be aggregated in the Lie algebra space under the first-order Baker-Campbell-Hausdorff (BCH) approximation.
Based on this observation, we introduce Temporally Coupled Pose Integration (TCPI), which adaptively aggregates neighboring trajectory states within a local temporal window into a temporally coupled pose for rendering.
Moreover, since each temporally coupled pose jointly depends on multiple neighboring trajectory locations, the reconstruction loss couples their updates and promotes temporally consistent pose refinement during joint pose-scene optimization.
We next present the trajectory integration formulation, the time-adaptive integration weighting, and numerical approximation.

\begin{figure*}
  \includegraphics[width=\textwidth]{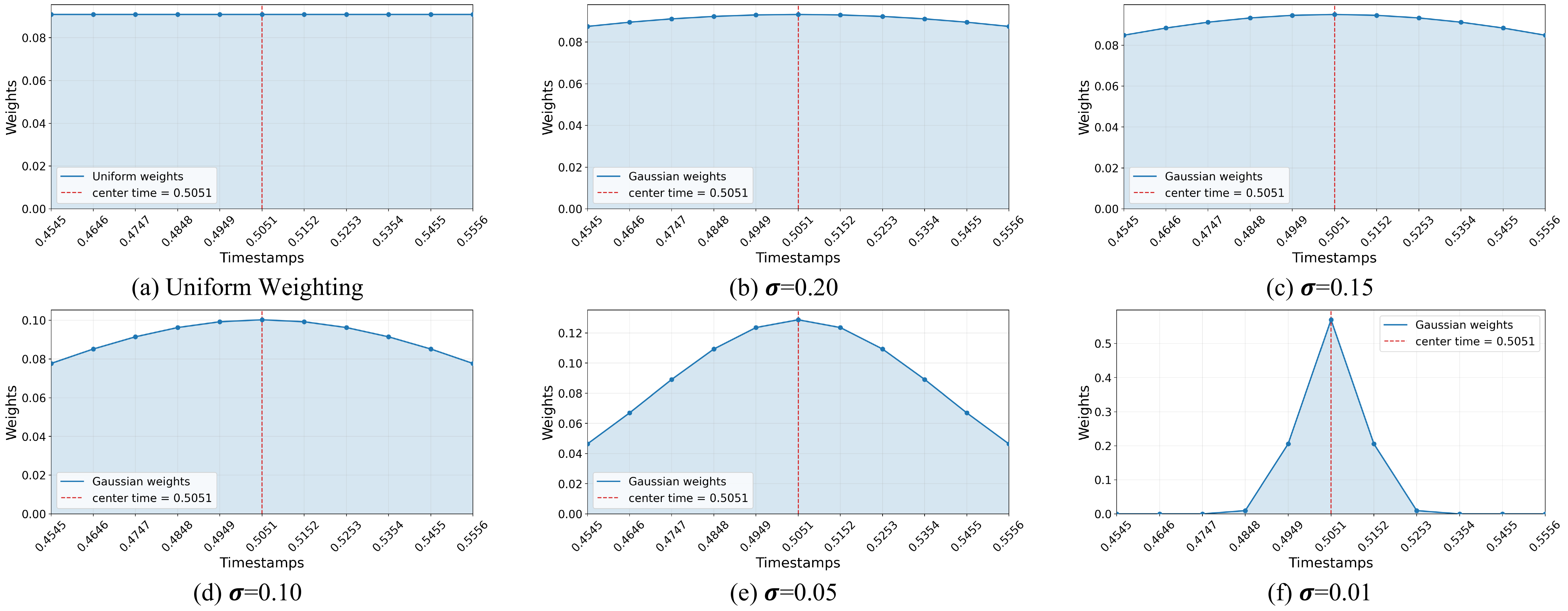}
  \caption{
    Effect of the temporal scale $\sigma$ on the integration weights within a local window centered at the query timestamp $\tau_k=0.5051$.
    \textbf{(a)} Uniform weighting assigns equal weights to all neighboring poses.
    \textbf{(b)}-\textbf{(f)} Truncated Gaussian weighting with $\sigma$ progressively decreasing from 0.2 to 0.01.
    A larger $\sigma$ produces a flatter distribution that approaches uniform weighting, whereas a smaller $\sigma$ concentrates the integration around $\tau_k$ and reduces the influence of distant poses.
  }
  \label{fig:distribution}
\end{figure*}

\noindent\textbf{Temporal Trajectory Integration.}
Given a query timestamp $\tau_k$, we query CTPF to obtain the corresponding camera pose $\mathbf{T}(\tau_k)\in\mathrm{SE}(3)$ and temporal scale $\sigma(\tau_k)$.
TCPI then constructs a temporally coupled pose $\tilde{\mathbf{T}}(\tau_k)$ by integrating the neighboring poses within the local temporal window $[\tau_k-\frac{\rho}{2},\tau_k+\frac{\rho}{2}]$ around the query timestamp.
Since camera poses lie on the nonlinear manifold $\text{SE}(3)$, they cannot be directly averaged in Euclidean space.
Under the BCH approximation, sufficiently small relative transformations within a local temporal window can be aggregated in the Lie algebra $\mathfrak{se}(3)$.
Let $\delta\in[-\frac{\rho}{2},\frac{\rho}{2}]$, taking $\mathbf{T}(\tau_k)$ as the reference pose, we map each neighboring pose to a relative increment as:
\begin{equation}
\Delta\boldsymbol{\xi}(\tau_k,\tau_k+\delta)=\operatorname{Log}\left(\mathbf{T}(\tau_k)^{-1}\mathbf{T}(\tau_k+\delta)\right),
\label{eq:relative_pose_increment}
\end{equation}
where $\operatorname{Log}$ denotes the Lie logarithm map from $\text{SE}(3)$ to $\mathfrak{se}(3)$.
By restricting the integration to a short temporal neighborhood, the relative transformations generally remain sufficiently small for local aggregation.
Temporally coupled pose is defined as:
\begin{equation}
\tilde{\mathbf{T}}(\tau_k)=\mathbf{T}(\tau_k)\operatorname{Exp}\left(\int_{-\frac{\rho}{2}}^{\frac{\rho}{2}}w(\tau_k,\tau_k\!+\!\delta)\Delta\boldsymbol{\xi}(\tau_k,\tau_k\!+\!\delta)\,d\delta\right),
\label{eq:integral}
\end{equation}
where $\operatorname{Exp}$ denotes the Lie exponential map from $\mathfrak{se}(3)$ to $\text{SE}(3)$, and
$w(\tau_k,\cdot)$ is a temporal integration weight.

\noindent\textbf{Time-Adaptive Integration Weighting.}
As shown in \cref{fig:distribution} \textbf{(a)}, uniform weighting assigns equal importance to all poses within the local temporal window.
It is effective when the camera motion is locally smooth, where neighboring poses provide comparable and reliable motion information.
However, distant poses may yield larger relative deviations from the query pose under rapid motion changes, whereas they should receive smaller contributions during integration.

To adapt to time-varying dynamic motion, we employ a truncated Gaussian weighting function, whose scale determines the effective temporal range of trajectory integration.
As illustrated in \cref{fig:distribution} \textbf{(b)}-\textbf{(f)}, a larger scale produces a flatter distribution that approaches uniform weighting, allowing more neighboring poses to contribute and preserving stronger temporal coupling when the local motion is smooth.
In contrast, a smaller scale concentrates the weights around the query timestamp, reducing the influence of distant poses and improving robustness to rapidly varying motion.
However, excessively reducing the scale weakens the temporal coupling among neighboring poses and gradually degenerates TCPI toward independent pose updates.

Therefore, a fixed scale imposes the same integration range at every timestamp and cannot adapt to the varying motion patterns along the trajectory.
We thus use the time-dependent scale $\sigma(\tau_k)$ predicted by CTPF in \cref{eq:bounded_sigma} to adapt the effective integration range at each timestamp.
Accordingly, the temporal integration weight is defined as a truncated Gaussian distribution centered at $\tau_k$:
\begin{equation}
w(\tau_k,\tau_k+\delta)
=
\frac{
\exp\left(
-\frac{\delta^2}{2\sigma(\tau_k)^2}
\right)
}{
\displaystyle
\int_{-\frac{\rho}{2}}^{\frac{\rho}{2}}
\exp\left(
-\frac{u^2}{2\sigma(\tau_k)^2}
\right)
\,du
},
\label{eq:truncated_gaussian}
\end{equation}
where the time-dependent scale $\sigma(\tau_k)$ allows TCPI to adapt the effective extent of local pose integration at each timestamp.

\noindent\textbf{Numerical Approximation.}
To numerically approximate \cref{eq:integral}, we first uniformly sample $M$ temporal offsets as:
\begin{equation}
\delta_i
\sim
\mathcal{U}
\left(
-\frac{\rho}{2},
\frac{\rho}{2}
\right),
\qquad
i=1,\ldots,M,
\end{equation}
where $\mathcal{U}$ denotes the uniform distribution.
Querying CTPF at $\{\tau_k+\delta_i\}_{i=1}^{M}$ yields the local pose set
$\{\mathbf{T}(\tau_k+\delta_i)\}_{i=1}^{M}$.
The temporally coupled pose is then approximated as:
\begin{equation}
\hat{\mathbf{T}}(\tau_k)
=
\mathbf{T}(\tau_k)
\operatorname{Exp}
\left(
\sum_{i=1}^{M}
\hat{w}_{k,i}
\Delta\boldsymbol{\xi}
(\tau_k,\tau_k+\delta_i)
\right),
\label{eq:numerical}
\end{equation}
where the normalized discrete weight is defined as:
\begin{equation}
\hat{w}_{k,i}
=
\frac{
w(\tau_k,\tau_k+\delta_i)
}{
\sum_{j=1}^{M}
w(\tau_k,\tau_k+\delta_j)
}.
\label{eq:discrete_weight}
\end{equation}

The resulting poses are then used in the event-based reconstruction objective introduced in the following subsection.
During joint optimization, the two endpoints of each sampled event interval are rendered using their corresponding temporally coupled poses.
Each temporally coupled pose depends on a local trajectory segment, such that the event reconstruction loss jointly supervises the neighboring trajectory locations around each endpoint rather than independently updating two isolated poses.
In this manner, TCPI promotes temporally consistent pose updates and stabilizes the joint optimization of the continuous camera trajectory and 3DGS representation.

\subsection{Event-Driven Joint Optimization}\label{sec:edjo}
With the temporally coupled poses constructed by TCPI, we jointly optimize the continuous trajectory and 3DGS representation directly from event measurements.
For each sampled event interval, TCPI constructs poses at its two endpoints, which are used to render the corresponding intensity images.
The predicted logarithmic intensity change between these images is then matched against that encoded by the accumulated events, allowing the same event-based objective to supervise both the CTPF and 3DGS parameters.

According to the event generation model introduced in \Cref{sec:preliminary}, an event is triggered when the logarithmic intensity change at a pixel reaches the contrast threshold $C$.
Therefore, the events accumulated over a temporal interval $[\tau_k,\tau_k']$, where $\tau_k'=\tau_k+\Delta\tau$, encode the corresponding logarithmic intensity change.
The event-based intensity change is expressed as:
\begin{equation}
\Delta \mathbf{L}_{x,y}(\tau_k,\tau_k')
=
\log
\frac{\mathbf{I}_{x,y}(\tau_k')}{\mathbf{I}_{x,y}(\tau_k)}
=
C
\!\!\!\!
\sum_{\tau\in[\tau_k,\tau_k']}
\!\!\!\!
p_{x,y}(\tau),
\label{eq:event_sum}
\end{equation}
where $p_{x,y}(\tau)\in\{-1,+1\}$ denotes the polarity of the event triggered at pixel position $(x,y)$ and timestamp $\tau$.

As illustrated in \cref{fig:overall} \textbf{(c)}, for a sampled event interval $[\tau_k,\tau_k']$, TCPI constructs the corresponding temporally coupled poses $\hat{\mathbf{T}}(\tau_k)$ and $\hat{\mathbf{T}}(\tau_k')$ according to \cref{eq:numerical}, respectively.
Using these two temporally coupled poses, we render two intensity images $\hat{\mathbf{I}}(\tau_k)$ and $\hat{\mathbf{I}}(\tau_k')$ from the 3DGS representation.
The logarithmic intensity change is then computed as:
\begin{equation}
\Delta \hat{\mathbf{L}}(\tau_k,\tau_k')
=
\log \hat{\mathbf{I}}(\tau_k')
-
\log \hat{\mathbf{I}}(\tau_k).
\label{eq:rendered_log_difference}
\end{equation}
We optimize the CTPF parameters $\Theta$ and the 3DGS parameters $\Phi$ by minimizing the event-driven reconstruction loss:
\begin{equation}
\begin{aligned}
\mathcal{L}_{\text{event}}
={}&
\lambda_{\text{ssim}}
\mathcal{L}_{\text{ssim}}
\left(
\Delta \hat{\mathbf{L}}(\tau_k,\tau_k'),
\Delta \mathbf{L}(\tau_k,\tau_k')
\right)
\\
&+
(1-\lambda_{\text{ssim}})
\mathcal{L}_{1}
\left(
\Delta \hat{\mathbf{L}}(\tau_k,\tau_k'),
\Delta \mathbf{L}(\tau_k,\tau_k')
\right),
\end{aligned}
\label{eq:event_gray}
\end{equation}
where $\lambda_{\text{ssim}}$ balances the structural-similarity and pixel-wise reconstruction terms and is empirically set to $0.2$.
Since $\Delta\hat{\mathbf{L}}(\tau_k,\tau_k')$ jointly depends on the temporally coupled poses generated by CTPF and TCPI and the intensity images rendered from the 3DGS representation, minimizing $\mathcal{L}_{\text{event}}$ simultaneously updates the CTPF parameters $\Theta$ and the 3DGS parameters $\Phi$.
In this manner, the event measurements jointly refine the continuous camera trajectory and scene representation within a unified optimization objective.

\begin{algorithm2e}[!t]
    \footnotesize
    \SetKwInOut{Input}{input}
    \SetKwInOut{Output}{output}
    \SetKwFor{With}{with}{do}{end}
    \caption{Overall optimization procedure of \textit{EvTrajGS} with LRES.}
    \label{alg:training}

    \KwIn{
        Event sequence $\mathcal{E}$, model parameters
        $(\Theta,\Phi)$, number of temporal intervals
        $N^{\mathrm{r}}$, numbers of iterations
        $K^{\mathrm{u}}$ and $K^{\mathrm{r}}$, and
        reweighting factor $\beta$
    }
    \KwOut{Optimized model parameters $(\Theta,\Phi)$.}

    \While{not converged}{
        \tcc{\emph{Uniform-sampling phase.}}
        \For{$k=1$ \KwTo $K^{\mathrm{u}}$}{
            $[\tau,\tau']
            \leftarrow
            \operatorname{UniformSample}(\mathcal{E})$\;

            $(\Theta,\Phi)
            \leftarrow
            \operatorname{Update}
            \bigl(
                (\Theta,\Phi),
                \mathcal{L}_\text{event}(\tau,\tau')
            \bigr)$\;
        }

        \tcc{\emph{LRES phase.}}
        $\{[\tau_i^{\mathrm{r}},\tau_{i+1}^{\mathrm{r}}]\}
        _{i=1}^{N^{\mathrm{r}}}
        \leftarrow
        \operatorname{Partition}
        ([\tau_{\min},\tau_{\max}],N^{\mathrm{r}})$\;
        
        \With{stop gradient}{
          Compute $\boldsymbol{\ell}\leftarrow\{\ell_i\}_{i=1}^{N^{\mathrm{r}}}$ using \cref{eq:lres_interval_loss}\;
          Compute $\boldsymbol{\pi}$ with $\beta$ using \cref{eq:loss_based_weight}\;
        }
        \For{$k=1$ \KwTo $K^{\mathrm{r}}$}{
            $j
            \leftarrow
            \operatorname{CategoricalSample}
            (\boldsymbol{\pi})$\;

            $[\tau,\tau']
            \leftarrow
            \operatorname{UniformSample}
            ([\tau_j^{\mathrm{r}},
              \tau_{j+1}^{\mathrm{r}}])$\;

            $(\Theta,\Phi)
            \leftarrow
            \operatorname{Update}
            \bigl(
                (\Theta,\Phi),
                \mathcal{L}_\text{event}(\tau,\tau')
            \bigr)$\;
        }
    }
\end{algorithm2e}

\subsection{Loss-Reweighted Event Sampling}\label{sec:lres}
Existing event-based reconstruction methods typically sample event intervals uniformly during optimization, assigning equal probability to all temporal regions regardless of their current fitting status.
However, different intervals may vary substantially in reconstruction difficulty.
Consequently, uniform sampling may repeatedly revisit well-fitted intervals, while more challenging intervals receive insufficient supervision.

To allocate optimization effort more effectively, we introduce Loss-Reweighted Event Sampling (LRES), which alternates between uniform and LRES phases.
The uniform-sampling phase maintains coverage over the entire event sequence, whereas the LRES phase allocates additional samples to intervals with larger reconstruction losses.
The overall optimization procedure is summarized in \cref{alg:training}.

After each uniform-sampling phase of $K^{\mathrm{u}}$ iterations, we uniformly partition the complete temporal span of the event stream into
$N^{\mathrm{r}}$ non-overlapping intervals $\{[\tau_i^{\mathrm{r}},\tau_{i+1}^{\mathrm{r}}]\}_{i=1}^{N^{\mathrm{r}}}$.
We evaluate the value of $\mathcal{L}_\text{event}$ in \cref{eq:event_gray} in each interval as:
\begin{equation}
\ell_i
=
\mathcal{L}_{\mathrm{event}}
\left(
\tau_i^{\mathrm{r}},
\tau_{i+1}^{\mathrm{r}}
\right),
\ 
i=1,\ldots,N^{\mathrm{r}},
\label{eq:lres_interval_loss}
\end{equation}
These loss values are evaluated without gradient propagation and used solely to form the subsequent sampling distribution.

Let $\boldsymbol{\ell}=\{\ell_i\}^{N^{\mathrm{r}}}_{i=1}$ denote the interval-loss vector.
The corresponding sampling probabilities are computed as:
\begin{equation}
\boldsymbol{\pi}
=
\operatorname{Softmax}
\left(
\beta\boldsymbol{\ell}
\right),
\label{eq:loss_based_weight}
\end{equation}
where $\boldsymbol{\pi}=\{\pi_i\}^{N^{\mathrm{r}}}_{i=1}$ and $\beta\in[0,2]$ controls the concentration of the sampling distribution.
When $\beta=0$, all intervals are assigned equal probabilities, which is equivalent to uniform sampling, whereas increasing $\beta$ progressively shifts the distribution toward intervals with larger loss values.

During the subsequent LRES phase of $K^{\mathrm{r}}$ iterations, event intervals are drawn from the categorical distribution defined by $\boldsymbol{\pi}$.
Sampling from this categorical distribution is implemented using inverse transform sampling (ITS).
Specifically, we first sample a temporal region according to $\boldsymbol{\pi}$.
An event interval is then uniformly sampled within the selected region and used to update the CTPF and 3DGS parameters under $\mathcal{L}_{\text{event}}$.
After $K^{\mathrm{r}}$ iterations, the optimization returns to the uniform-sampling phase, and the interval losses are periodically recomputed to reflect the current fitting status.
This alternating strategy maintains global coverage of the event sequence while allocating additional optimization effort to underfitted intervals, improving reconstruction fidelity.

\section{Experiments}
\subsection{Experimental Setups}
\noindent\textbf{Datasets.}
We evaluate the reconstruction quality and pose accuracy of \textit{EvTrajGS} on both synthetic and real-world datasets.

\textit{For the synthetic dataset,}
we follow the experimental settings of IncEventGS \cite{2025inceventgs} and generate event data using 3D scene models from the Replica dataset \cite{2019replica}.
We select five indoor scenes: \textsl{Room0}, \textsl{Room2}, \textsl{Office0}, \textsl{Office2}, and \textsl{Office3}.
Each scene is rendered along the camera trajectory used in NICE-SLAM \cite{zhu2022nice-slam} to produce a high-frame-rate RGB sequence at 1000 Hz with a spatial resolution of $768\times480$.
The resulting RGB sequences serve as the ground-truth images.
The corresponding event streams are synthesized from the rendered images using the event simulator proposed in \cite{gehrig2020video}.

\textit{For the real-world dataset,}
we follow the experimental settings of IncEventGS \cite{2025inceventgs} and conduct evaluations on the TUM-VIE dataset \cite{2021tum_vie}.
TUM-VIE captures event streams using a stereo pair of Prophesee Gen4 HD event cameras with a spatial resolution of $1280\times720$.
Following prior work, we use only the event streams captured by the left camera and select five indoor sequences for evaluation: \textsl{1d}, \textsl{3d}, \textsl{6dof}, \textsl{Desk}, and \textsl{Desk2}.
Since TUM-VIE does not provide ground-truth images aligned with the event camera, we choose the closest images from the RGB camera and crop them to the same size as the rendered images for visual comparisons.

\noindent\textbf{Baselines.} We compare \textit{EvTrajGS} against six state-of-the-art methods and three representative two-stage pipelines.
The state-of-the-art methods include three NeRF-based reconstruction approaches: E-NeRF \cite{2023e_nerf}, EventNeRF \cite{2023eventnerf}, and Robust e-NeRF \cite{2023robust_e_nerf}, as well as a fixed-pose 3DGS reconstruction method: EventSplat$^\dagger$ \cite{yura2025eventsplat}, and two SLAM-style joint optimization methods: IncEventGS \cite{2025inceventgs} and E2EGS \cite{kim2026e2egs}.
We further construct three two-stage pipelines by combining E2VID \cite{2019e2vid} and off-the-shelf pose estimation methods, including COLMAP \cite{2016colmap} and VGGT \cite{wang2025vggt}, with image-based ($\mathcal{L}_{\text{image}}$) and event-based ($\mathcal{L}_{\text{event}}$) reconstruction losses in 3DGS \cite{3dgs}.
Following the experimental protocols in \cite{2025inceventgs}, the NeRF-based methods are trained on event streams using ground-truth camera poses.
Since the source code of EventSplat and E2EGS is unavailable, we reproduce EventSplat$^\dagger$ following the E2VID+COLMAP+$\mathcal{L}_{\text{event}}$ pipeline described in its paper, while reporting the E2EGS results directly from its paper.

\begin{figure}
\centering
\includegraphics[width=\columnwidth]{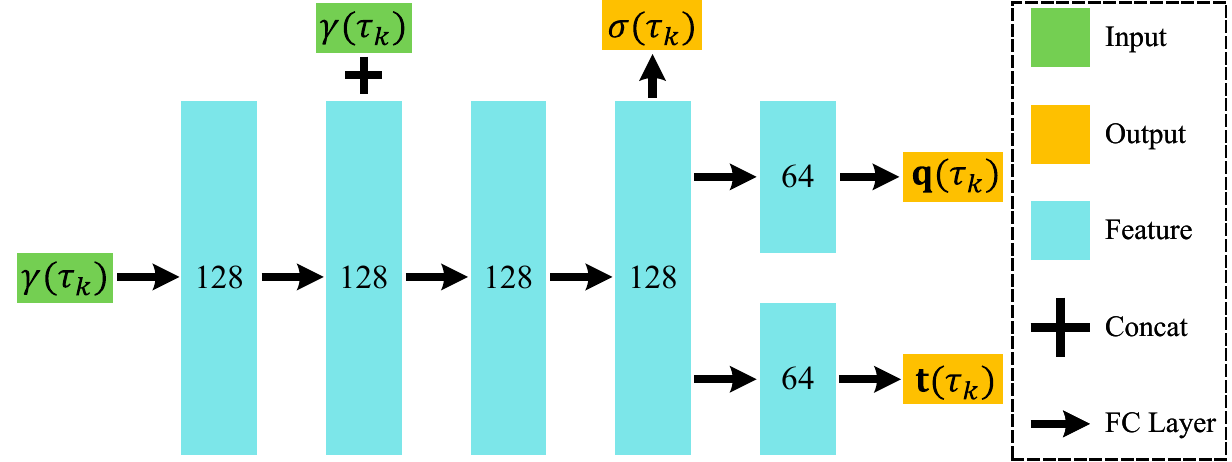}
\caption{
  The architecture of the multi-layer perceptron (MLP) used for the proposed CTPF.
  Given a timestamp $\tau_k$, we first encode it using the Fourier positional encoding $\gamma(\cdot)$ in \cref{eq:PE}.
  The resulting feature is then processed by four fully connected (FC) layers, each with 128 channels and a ReLU activation.
  A skip connection concatenates the encoded input feature with the output of the second hidden layer.
  From the resulting latent feature, the network predicts the time-dependent scale $\sigma(\tau_k)\in\mathbb{R}$ through \cref{eq:bounded_sigma}.
  Finally, the latent feature is fed into two separate FC branches, each reducing the feature dimension to 64 channels, to predict the rotation and translation, respectively.
  The rotation is represented by a unit quaternion $\mathbf{q}(\tau_k)\in\mathbb{R}^4$ obtained by \cref{eq:quaternion_normalization}, while the translation is represented by $\mathbf{t}(\tau_k)\in\mathbb{R}^3$.
}
\label{fig:mlp}
\end{figure}

\noindent\textbf{Evaluation Metrics.}
We evaluate \textit{EvTrajGS} and the baseline methods from two aspects:
\textbf{(1) Reconstruction Quality:}
We measure the visual similarity between rendered and ground-truth images using Peak Signal-to-Noise Ratio (PSNR), Structural Similarity Index Measure (SSIM) \cite{ssim}, and Learned Perceptual Image Patch Similarity (LPIPS) \cite{lpips}.
For fair comparison, we use the official evaluation code of IncEventGS \cite{2025inceventgs}, which applies a linear color transformation to the rendered images before metric computation.
\textbf{(2) Pose Accuracy:}
We evaluate camera pose accuracy using the Absolute Trajectory Error (ATE), which measures the translational discrepancy between the estimated and reference trajectories after trajectory alignment.
The reported metric is the root mean square error (RMSE) of ATE in centimeters, computed using the publicly available EVO toolbox \cite{grupp2017evo}.
For fair comparison, we recompute all ATE results presented in the IncEventGS paper.\footnote{The IncEventGS authors confirmed an error in their EVO evaluation command; see the corresponding \href{https://github.com/WU-CVGL/IncEventGS/issues/13}{GitHub issue}.}

\begin{table*}[!t]
\caption{
Quantitative comparison of novel-view synthesis on the Replica dataset \cite{2019replica}.
We report PSNR, SSIM, and LPIPS for five scenes and their averages.
\textbf{Bold} text indicates the best performance.
$^\dagger$ denotes our re-implementation based on the original paper.
}
\label{tab:Replica}
\centering
\footnotesize
\setlength{\tabcolsep}{1mm}
\resizebox{\textwidth}{!}{
\begin{tabular}{lccccccccccccccccccccc!{\vrule width 0.8pt}ccc}
\toprule
\multirow{2}{*}{\textbf{Method}}
&
\multirow{2}{*}{\textbf{Venue}}
&
\multicolumn{3}{c}{\textsl{Room0}}
&&
\multicolumn{3}{c}{\textsl{Room2}}
&&
\multicolumn{3}{c}{\textsl{Office0}}
&&
\multicolumn{3}{c}{\textsl{Office2}}
&&
\multicolumn{3}{c}{\textsl{Office3}}
&&
\multicolumn{3}{c}{\textbf{Average}}
\\
\cmidrule{3-5}
\cmidrule{7-9}
\cmidrule{11-13}
\cmidrule{15-17}
\cmidrule{19-21}
\cmidrule{23-25}
&
&
PSNR$\uparrow$ & SSIM$\uparrow$ & LPIPS$\downarrow$
&&
PSNR$\uparrow$ & SSIM$\uparrow$ & LPIPS$\downarrow$
&&
PSNR$\uparrow$ & SSIM$\uparrow$ & LPIPS$\downarrow$
&&
PSNR$\uparrow$ & SSIM$\uparrow$ & LPIPS$\downarrow$
&&
PSNR$\uparrow$ & SSIM$\uparrow$ & LPIPS$\downarrow$
&&
PSNR$\uparrow$ & SSIM$\uparrow$ & LPIPS$\downarrow$
\\
\midrule

\multicolumn{20}{l}{\textsl{NeRF-based reconstruction methods}} \\

E-NeRF \cite{2023e_nerf}
& RA-L'23
& 13.99 & 0.58 & 0.51
&& 15.56 & 0.47 & 0.58
&& 18.91 & 0.51 & 0.57
&& 13.05 & 0.65 & 0.44
&& 14.01 & 0.62 & 0.48
&& 15.10 & 0.57 & 0.52
\\

EventNeRF \cite{2023eventnerf}
& CVPR'23
& 17.29 & 0.62 & 0.39
&& 16.02 & 0.54 & 0.64
&& 18.90 & 0.43 & 0.62
&& 15.18 & 0.66 & 0.45
&& 16.77 & 0.73 & 0.33
&& 16.83 & 0.60 & 0.49
\\

Robust e-NeRF \cite{2023robust_e_nerf}
& ICCV'23
& 17.26 & 0.84 & 0.18
&& 16.43 & 0.50 & 0.52
&& 18.93 & 0.52 & 0.56
&& 16.81 & 0.81 & 0.25
&& 19.22 & 0.84 & 0.18
&& 17.73 & 0.70 & 0.34
\\

\midrule
\multicolumn{20}{l}{\textsl{Fixed-pose 3DGS reconstruction methods}} \\

E2VID + COLMAP + $\mathcal{L}_{\text{image}}$
& --
& 16.48 & 0.74 & 0.30
&& 17.95 & 0.72 & 0.27
&& 21.18 & 0.52 & 0.44
&& 16.30 & 0.74 & 0.33
&& 17.83 & 0.76 & 0.26
&& 17.95 & 0.70 & 0.32
\\

E2VID + VGGT + $\mathcal{L}_{\text{image}}$
& --
& 16.60 & 0.75 & 0.28
&& 18.06 & 0.75 & 0.27
&& 20.30 & 0.51 & 0.49
&& 16.12 & 0.73 & 0.30
&& 17.73 & 0.74 & 0.31
&& 17.76 & 0.70 & 0.33
\\

E2VID + VGGT + $\mathcal{L}_{\text{event}}$
& --
& 17.43 & 0.74 & 0.21
&& 21.72 & 0.78 & 0.22
&& 23.65 & 0.55 & 0.37
&& 18.53 & 0.75 & 0.29
&& 18.96 & 0.79 & 0.19
&& 20.06 & 0.72 & 0.26
\\

EventSplat$^{\dagger}$ \cite{yura2025eventsplat}
& CVPR'25
& 21.36 & 0.75 & 0.22
&& 22.36 & 0.76 & 0.26
&& 23.89 & 0.57 & 0.35
&& 19.59 & 0.78 & 0.27
&& 20.67 & 0.83 & 0.17
&& 21.57 & 0.74 & 0.25
\\\midrule

\multicolumn{20}{l}{\textsl{SLAM-style joint optimization methods}} \\
IncEventGS \cite{2025inceventgs}
& CVPR'25
& 24.31 & 0.85 & 0.17
&& 23.75 & 0.79 & 0.23
&& 25.64 & 0.54 & 0.30
&& 21.74 & 0.82 & 0.23
&& 21.18 & 0.88 & 0.13
&& 23.32 & 0.78 & 0.21
\\

E2EGS \cite{kim2026e2egs}
& CVPR'26
& 23.86 & 0.87 & 0.19
&& 23.01 & 0.77 & 0.26
&& 28.01 & 0.52 & 0.41
&& 24.86 & 0.83 & 0.25
&& 20.75 & 0.85 & 0.19
&& 24.10 & 0.77 & 0.26
\\\midrule

\rowcolor{MyGray}
\textbf{EvTrajGS (Ours)}
& --
& \textbf{28.59} & \textbf{0.91} & \textbf{0.10}
&& \textbf{26.17} & \textbf{0.92} & \textbf{0.10}
&& \textbf{28.51} & \textbf{0.74} & \textbf{0.15}
&& \textbf{26.59} & \textbf{0.90} & \textbf{0.13}
&& \textbf{29.83} & \textbf{0.92} & \textbf{0.09}
&& \textbf{27.94} & \textbf{0.88} & \textbf{0.11}
\\

\bottomrule
\end{tabular}
}
\end{table*}

\begin{figure*}
\centering
\includegraphics[width=\textwidth]{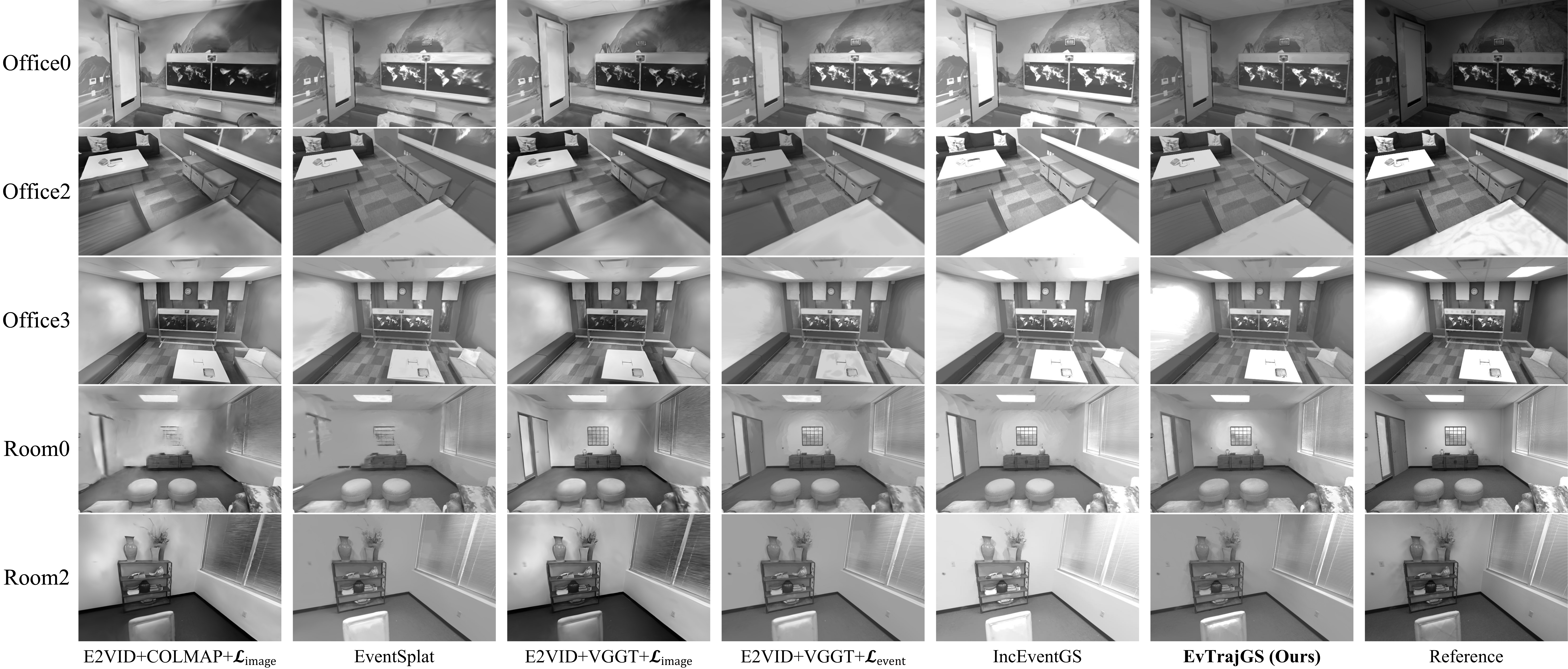}
\caption{Qualitative comparisons with state-of-the-art methods on the synthetic Replica \cite{2019replica} dataset.}
\label{fig:visual_replica}
\end{figure*}

\noindent\textbf{Implementation Details.}
\textit{EvTrajGS} is implemented on top of GSplat \cite{ye2025gsplat}, a PyTorch-based \cite{Pytorch} implementation of 3DGS.
All experiments conducted by us were run on a single NVIDIA RTX 3090 GPU.
The complete pipeline takes approximately $20$ minutes, including $3$ minutes for E2VID reconstruction, $1$ minute for VGGT pose estimation, $1$ minute for CTPF initialization, and $15$ minutes for subsequent joint optimization.

\textit{For CTPF,}
we reconstruct intensity frames from the event stream at intervals of $50$ ms using E2VID \cite{2019e2vid}.
The resulting frame sequence is then processed by VGGT \cite{wang2025vggt} to estimate coarse camera poses and initial 3D points.
CTPF is parameterized by a lightweight MLP $F_\Theta$, whose architecture is illustrated in \cref{fig:mlp}.
The $\omega$ in the Fourier positional encoding is set to 4 by default.
It is initialized in a full-batch manner for $2000$ iterations using Adam \cite{adam} with a weight decay of $1\times10^{-6}$.
The learning rate is cosine annealed from $2\times10^{-3}$ to $2\times10^{-4}$.
After initialization, $\mathcal{L}_{\mathrm{pose}}$ is removed, and CTPF is refined solely under the event-based objective.

\textit{For TCPI,}
all temporal hyperparameters are defined in the normalized time domain.
We pad the temporal range by 10\% of the original duration, yielding a normalized padding ratio of $\frac{1}{11}$ \textit{w.r.t} the padded range, while $\sigma_{\min}$ and $\sigma_{\max}$ are set to $0.01$ and $0.2$, respectively.
The number of temporal samples $M$ is set to $11$ for Replica and $5$ for TUM-VIE.

\textit{For joint optimization,}
the learning rates and optimization schedule of the 3DGS parameters follow the default GSplat configuration.
The learning rate of CTPF is cosine annealed from $5\times10^{-6}$ to $2\times10^{-6}$.
The total number of optimization iterations is set to $15000$ using Adam \cite{adam} with a weight decay of $1\times10^{-6}$.
The contrast threshold $C$ of Replica and TUM-VIE is set to $0.1$ and $0.2$, respectively.
$\beta$ is set to $1$, while $K^{\mathrm{u}}$ and $K^{\mathrm{r}}$ are both set to $1000$ iterations.
The complete temporal span is partitioned into $N^{\mathrm{r}}$$=$$100$ intervals.
Each distribution update takes only $1$$\sim$$2$ seconds and introduces a minor cost.

\begin{table*}[!t]
\caption{
Quantitative comparison of camera pose accuracy on the synthetic Replica \cite{2019replica} and real-world TUM-VIE \cite{2021tum_vie} datasets.
We report ATE RMSE$\downarrow$ in centimeters, computed using the EVO toolkit \cite{grupp2017evo}.
\textbf{Bold} text indicates the best performance.
}
\label{tab:pose_acc}
\footnotesize
\centering
\resizebox{\textwidth}{!}{
\begin{tabular}{lcccccc!{\vrule width 0.8pt}ccccccc!{\vrule width 0.8pt}c}
\toprule
\multirow{2}{*}{\textbf{Method}}
& \multirow{2}{*}{\textbf{Venue}}
& \multicolumn{6}{c}{\textbf{Replica} \cite{2019replica}}
&& \multicolumn{6}{c}{\textbf{TUM-VIE} \cite{2021tum_vie}} \\
\cmidrule(lr){3-8}
\cmidrule(lr){10-15}
&
& \textsl{Room0}
& \textsl{Room2}
& \textsl{Office0}
& \textsl{Office2}
& \textsl{Office3}
& \textbf{Avg.}
&
& \textsl{1d}
& \textsl{3d}
& \textsl{6dof}
& \textsl{Desk}
& \textsl{Desk2}
& \textbf{Avg.} \\
\midrule

\multicolumn{15}{l}{\textsl{Off-the-shelf pose estimation pipelines}} \\
E2VID \cite{2019e2vid} + COLMAP \cite{2016colmap}
& --
& 0.459 & 1.299 & 0.721 & 0.331 & 0.767 & 0.715
&& 0.138 & 0.471 & 0.261 & 2.331 & 0.719 & 0.784 \\

E2VID \cite{2019e2vid} + VGGT \cite{wang2025vggt}
& --
& 1.082 & 1.224 & 1.591 & 1.060 & 1.027 & 1.197
&& 0.258 & 1.047 & 1.339 & 5.165 & 1.133 & 1.788 \\

DEVO \cite{2024devo}
& 3DV'24
& 1.020 & 0.363 & 0.556 & 0.499 & 0.810 & 0.650
&& 0.158 & 2.400 & 0.365 & 0.964 & 1.314 & 1.040 \\\midrule

\multicolumn{15}{l}{\textsl{SLAM-style joint optimization methods}} \\
IncEventGS \cite{2025inceventgs}
&CVPR'25
& 0.222 & 0.138 & 0.272 & 0.215 & 0.280 & 0.225
&& 0.200 & 0.298 & 0.300 & 0.293 & 0.390 & 0.296 \\\midrule

\rowcolor{MyGray}
\textbf{EvTrajGS (Ours)}
&--
& \textbf{0.110}
& \textbf{0.110}
& \textbf{0.144}
& \textbf{0.068}
& \textbf{0.211}
& \textbf{0.129}
&& \textbf{0.105}
& \textbf{0.113}
& \textbf{0.221}
& \textbf{0.148}
& \textbf{0.178}
& \textbf{0.153} \\
\bottomrule
\end{tabular}
}
\end{table*}

\begin{figure*}
\centering
\includegraphics[width=\textwidth]{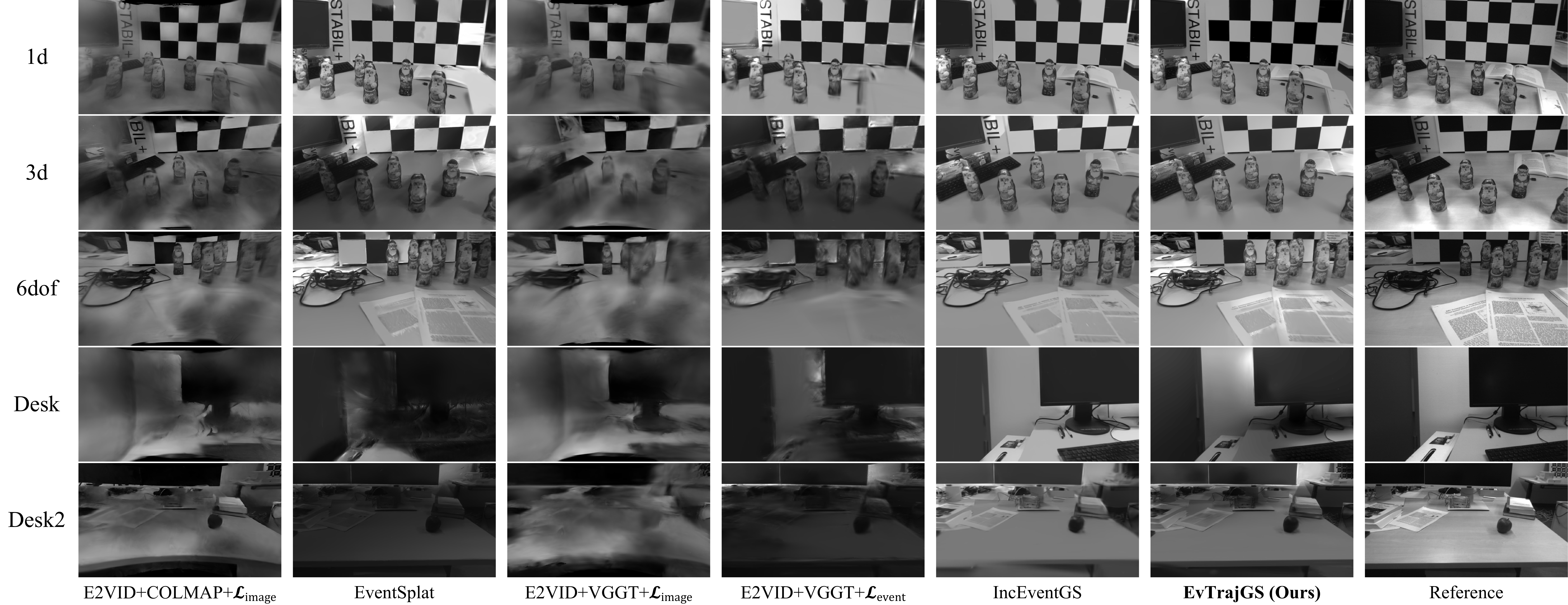}
\caption{Qualitative comparisons with state-of-the-art methods on the real-world TUM-VIE \cite{2021tum_vie} dataset.}
\label{fig:visual_tumvie}
\end{figure*}

\subsection{Experimental Results}
We evaluate \textit{EvTrajGS} from three aspects: reconstruction quality, pose accuracy, and computational efficiency.
These evaluations examine whether \textit{EvTrajGS} effectively addresses the accuracy-efficiency trade-off discussed in \cref{sec:intro}.

\noindent\textbf{Reconstruction Quality.}
We quantitatively evaluate novel-view synthesis on the Replica dataset and provide qualitative comparisons on both Replica and TUM-VIE datasets.

As reported in \Cref{tab:Replica}, \textit{EvTrajGS} achieves the best reconstruction quality across all five Replica scenes in terms of PSNR, SSIM, and LPIPS.
Compared with the second-best average result for each metric, it improves PSNR and SSIM by $3.84$ dB and $0.10$, respectively, while reducing LPIPS by $0.10$.
Notably, \textit{EvTrajGS} also consistently outperforms the NeRF-based methods evaluated with ground-truth camera poses, despite requiring only coarse pose initialization from an off-the-shelf pipeline.
The comparison with the fixed-pose E2VID+VGGT+$\mathcal{L}_{\text{event}}$ pipeline further highlights the importance of pose refinement.
Starting from the same pose estimates, \textit{EvTrajGS} improves the average PSNR from $20.06$ dB to $27.94$ dB.
This result is consistent with our motivation that local errors in off-the-shelf trajectories limit reconstruction fidelity when the estimated poses remain fixed.

Qualitative results on the Replica and TUM-VIE are shown in \cref{fig:visual_replica,fig:visual_tumvie}, respectively.
As shown, \textit{EvTrajGS} produces higher-fidelity reconstructions than the compared methods, with finer details, clearer structures, and fewer visible artifacts.
These results are consistent with the quantitative improvements and further validate the effectiveness of our designs.

\noindent\textbf{Pose Accuracy.}
We evaluate pose accuracy on the Replica and TUM-VIE via ATE RMSE computed by the EVO toolkit \cite{grupp2017evo}.

As reported in \Cref{tab:pose_acc}, \textit{EvTrajGS} achieves the lowest ATE RMSE on every evaluated sequence.
Compared with IncEventGS, \textit{EvTrajGS} reduces the average ATE RMSE by $42.7\%$ on the Replica and $48.3\%$ on the TUM-VIE.
Moreover, relative to its E2VID+VGGT initialization, \textit{EvTrajGS} reduces the average ATE RMSE from $1.197$ cm to $0.129$ cm on the Replica and from $1.788$ cm to $0.153$ cm on the TUM-VIE.
These substantial improvements demonstrate that the proposed joint optimization can reliably correct the local inaccuracies in the coarse initial trajectories.

\begin{figure*}[!t]
  \centering
  \includegraphics[width=0.85\textwidth]{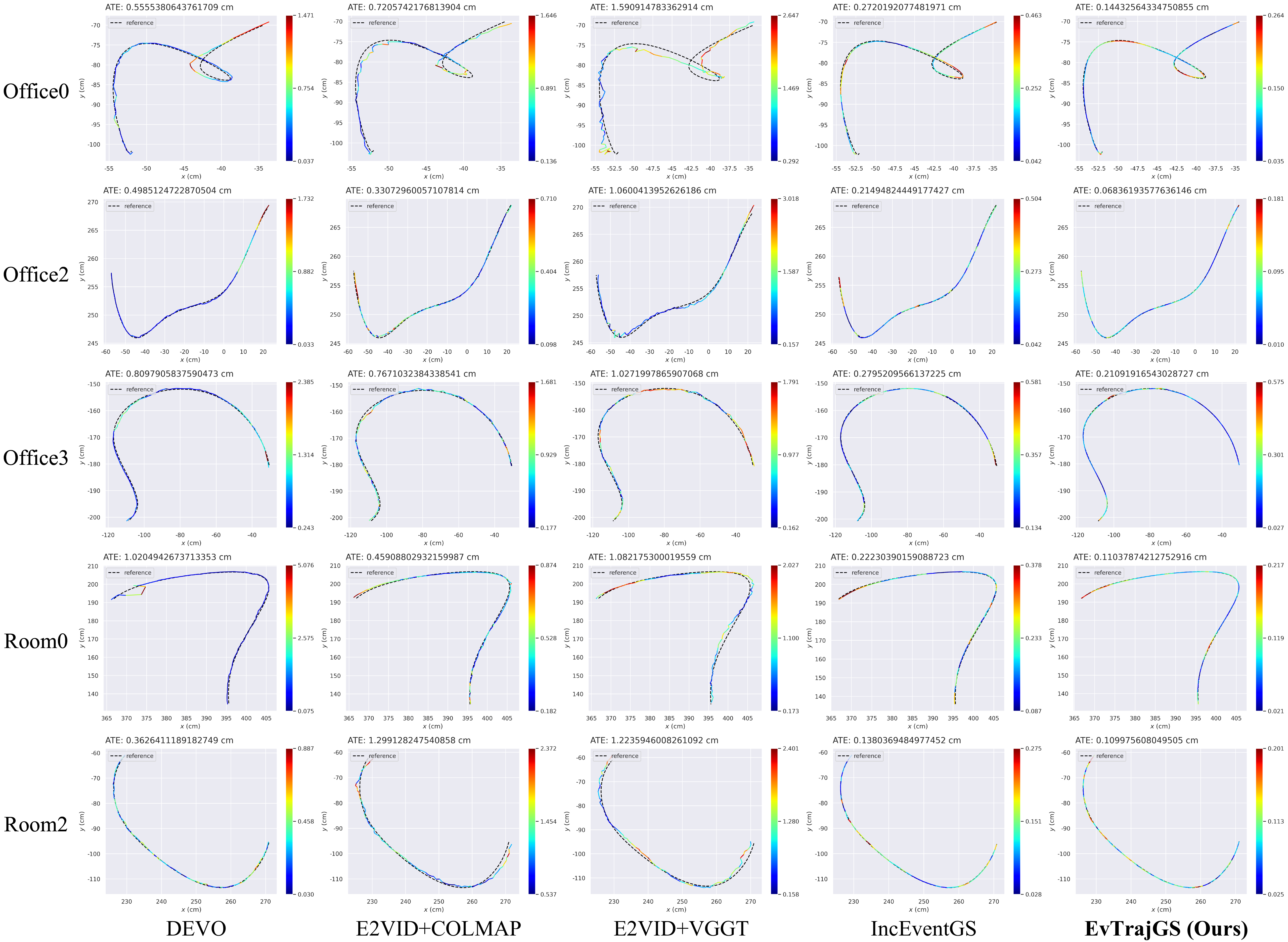}
  \caption{ATE visualizations on the Replica \cite{2019replica} dataset generated using the EVO toolbox \cite{grupp2017evo}. For each scene, all estimated trajectories are aligned to the same reference trajectory (dotted line). 
  Colors indicate point-wise trajectory errors using the individual range shown in each subplot.
  }
  \label{fig:pose_replica}
\end{figure*}

\begin{figure*}[!t]
  \centering
  \includegraphics[width=0.85\textwidth]{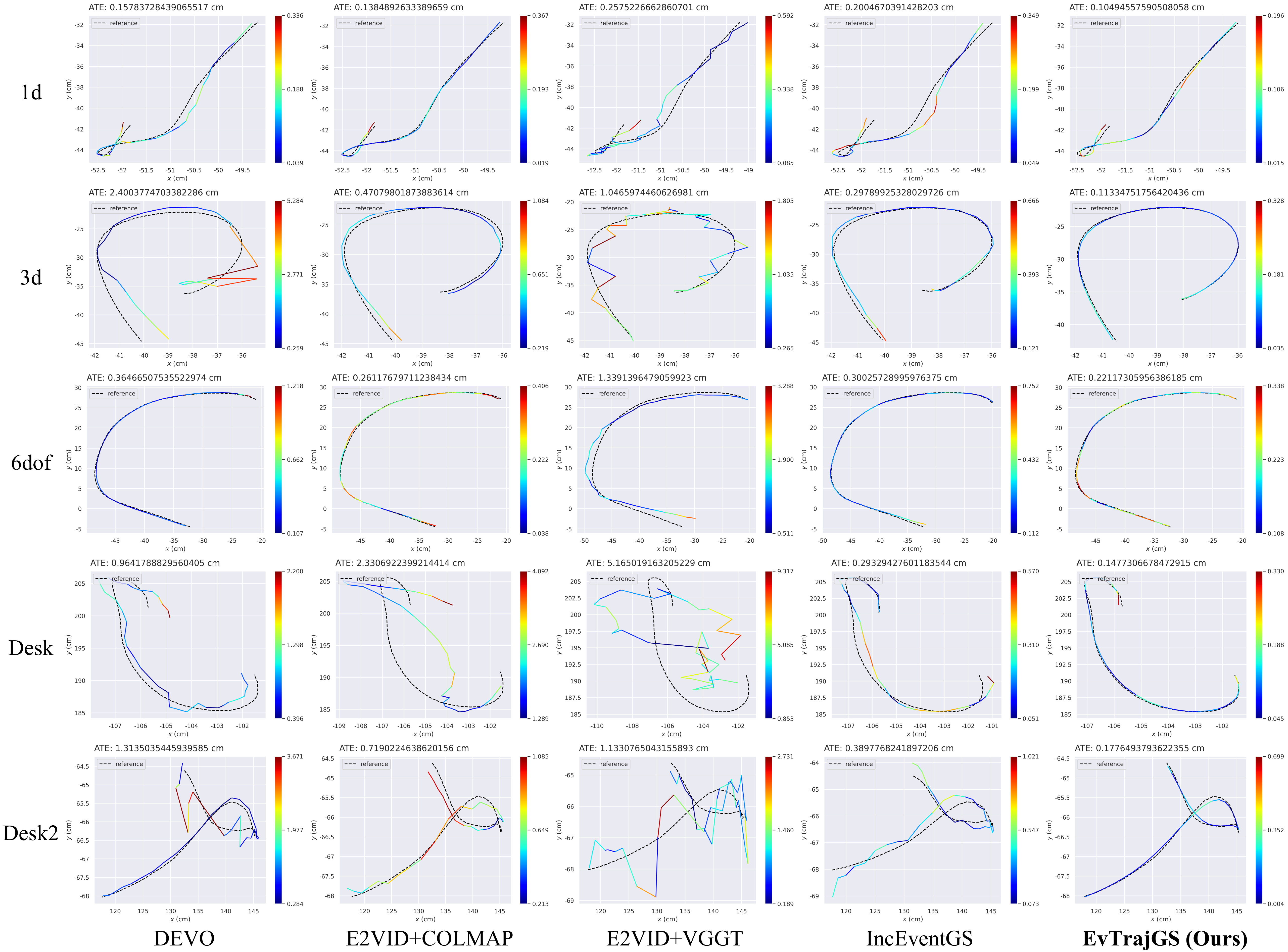}
  \caption{ATE visualizations on the TUM-VIE \cite{2021tum_vie} dataset generated using the EVO toolbox \cite{grupp2017evo}. For each scene, all estimated trajectories are aligned to the same reference trajectory (dotted line). 
  Colors indicate point-wise trajectory errors using the individual range shown in each subplot.}
  \label{fig:pose_tumvie}
\end{figure*}

We provide the ATE visualizations on the Replica and TUM-VIE datasets in \cref{fig:pose_replica,fig:pose_tumvie}, respectively.
The visualization results provide consistent qualitative evidence.
The trajectories estimated by \textit{EvTrajGS} remain more closely aligned with the reference ones.
By parameterizing camera motion as a shared continuous trajectory and constructing each rendering pose from a local trajectory segment, CTPF and TCPI provide coordinated trajectory refinement rather than independently updating isolated pose variables.

\noindent\textbf{Computational Efficiency.}
As shown in \cref{fig:time}, the complete \textit{EvTrajGS} pipeline takes approximately $20$ minutes on a single NVIDIA RTX 3090 GPU, including $5$ minutes for initialization and $15$ minutes for joint optimization.
In comparison, IncEventGS requires approximately $136$ minutes due to its repeated incremental tracking-and-mapping procedure and multi-view rendering during dense BA.
Although fixed-pose pipelines are computationally efficient, their reconstruction and pose accuracy remain substantially lower because the initial trajectory errors cannot be corrected.
\textit{EvTrajGS} breaks this trade-off by achieving the best reconstruction quality and pose accuracy without relying on costly mechanisms.

\subsection{Ablation Study and Parameter Sensitivity Analysis}
We conduct a series of experiments to analyze four aspects of \textit{EvTrajGS}:
\textbf{(1)} the contribution of each proposed component;
\textbf{(2)} the effectiveness and efficiency of the proposed CTPF+TCPI compared with various pose stabilization strategies;
\textbf{(3)} the influence of temporal weighting and scale selection in TCPI; and
\textbf{(4)} the sensitivity of LRES to the reweighting factor $\beta$.
All experiments are conducted on the Replica dataset \cite{2019replica} using a single NVIDIA RTX 3090 GPU.

\noindent\textbf{Ablation of Individual Components.}
We evaluate the contribution of each component in \Cref{tab:ablations}, where the baseline is set to E2VID+VGGT+$\mathcal{L}_{\text{image}}$ with fixed camera poses (row 1).
Replacing $\mathcal{L}_{\text{image}}$ with $\mathcal{L}_{\text{event}}$ (row 2) improves reconstruction quality, but the remaining pose errors continue to limit performance.
Direct joint optimization (row 3) further improves both reconstruction and pose accuracy, but its refinement remains limited when discrete poses are optimized independently.
Using CTPF (row 4) improves both reconstruction and pose accuracy, reducing ATE RMSE from $0.707$ cm to $0.497$ cm.
Adding TCPI (row 5) further reduces it to $0.144$ cm while consistently improving all reconstruction metrics, demonstrating the effectiveness of local trajectory integration.
Finally, LRES (row 6) further improves the results by emphasizing underfitted event intervals.
Together, these components progressively reduce the ATE RMSE from 1.197 cm to 0.129 cm and increase the PSNR from 17.76 dB to 27.94 dB.

\begin{table}[!t]
  \caption{
  Ablation study on the Replica dataset \cite{2019replica}.
  The first row denotes the baseline using E2VID+VGGT+$\mathcal{L}_\text{image}$, while each subsequent row incrementally introduces the corresponding component.
  \textbf{Bold} text indicates the best average performance.
  }
  \label{tab:ablations}
  \centering
  \setlength{\tabcolsep}{1mm}
  \resizebox{\columnwidth}{!}{
  \begin{tabular}{ccccc|cccc}
  \toprule
  $\mathcal{L}_{\text{event}}$
  & \textbf{Joint Opt.}
  & \textbf{CTPF}
  & \textbf{TCPI}
  & \textbf{LRES}
  & \textbf{PSNR$\uparrow$}
  & \textbf{SSIM$\uparrow$}
  & \textbf{LPIPS$\downarrow$}
  & \textbf{ATE RMSE$\downarrow$} \\
  \midrule

  & & & & 
  & 17.76 & 0.70 & 0.33 & 1.197 \\

  \ding{52}
  & & & &
  & 20.06 & 0.72 & 0.26 & 1.197 \\

  \ding{52}
  & \ding{52}
  & & &
  & 22.22 & 0.74 & 0.23 & 0.707 \\

  \ding{52}
  & \ding{52}
  & \ding{52}
  & &
  & 24.10 & 0.77 & 0.20 & 0.497 \\

  \ding{52}
  & \ding{52}
  & \ding{52}
  & \ding{52}
  &
  & 27.08 & 0.86 & 0.13 & 0.144 \\

  \rowcolor{MyGray}
  \ding{52}
  & \ding{52}
  & \ding{52}
  & \ding{52}
  & \ding{52}
  & \textbf{27.94}
  & \textbf{0.88}
  & \textbf{0.11}
  & \textbf{0.129} \\
  \bottomrule
  \end{tabular}
  }
\end{table}

\noindent\textbf{Different Pose Stabilization Strategies.}
We compare against different pose stabilization strategies under the same event-driven joint optimization framework.
All variants start from the E2VID+VGGT initialization and are evaluated on a single NVIDIA RTX 3090 GPU.
For first- and second-order pose regularization, the corresponding smoothness term is added to $\mathcal{L}_{\text{event}}$ with a coefficient of $0.05$.
For dense BA, the window size is set to $16$ as larger windows exceed the GPU memory.

\begin{table}[!t]
\caption{
Comparison of different pose stabilization strategies initialized from E2VID \cite{2019e2vid} + VGGT \cite{wang2025vggt} on the Replica dataset \cite{2019replica}.
All variants are optimized within the proposed event-driven joint optimization framework.
The reported time is measured on a single NVIDIA RTX 3090 GPU.
\textbf{Bold} text indicates the best performance.
}

\label{tab:pose_stabilization}
\setlength{\tabcolsep}{1mm}
\resizebox{\columnwidth}{!}{
\begin{tabular}{lccccc}
\toprule
\textbf{Strategy} & \textbf{PSNR$\uparrow$} & \textbf{SSIM$\uparrow$} & \textbf{LPIPS$\downarrow$} & \textbf{ATE RMSE$\downarrow$} & \textbf{Time}\\
\midrule
Direct Joint Opt.
&22.22 
&0.74 
&0.23 
&0.707 
&13 minutes\\
1st-order Pose Diff.
&22.26
&0.75
&0.22
&0.651
&13 minutes\\
2nd-order Pose Diff.
&24.19
&0.81
&0.17
&0.386
&13 minutes
\\
dense BA
&26.59
&0.84
&0.14
&0.182
&150 minutes\\
\rowcolor{MyGray}
CTPF+TCPI
&\textbf{27.08}
&\textbf{0.86}
&\textbf{0.13}
&\textbf{0.144}
&15 minutes\\
\bottomrule
\end{tabular}
}
\end{table}

As shown in \Cref{tab:pose_stabilization}, direct joint optimization provides limited pose refinement, while explicit pose difference regularization improves the results, with the second-order variant performing better than the first-order one.
Dense BA achieves strong performance but requires $150$ minutes and is limited by the memory overhead of sliding-window optimization.
In contrast, CTPF and TCPI achieve the best reconstruction and pose accuracy within only $15$ minutes by representing camera motion as a shared continuous trajectory and integrating local trajectory segments into temporally coupled poses.
These results demonstrate that our design effectively stabilizes pose updates at lower computational cost than dense BA.

\begin{table}[!t]
  \caption{
  Comparison of different temporal weighting and scale-selection strategies in TCPI on the Replica \cite{2019replica} dataset.
  \textbf{Bold} text indicates the best performance.
  }
  \label{tab:tcpi}
  \centering
  \resizebox{\columnwidth}{!}{
  \begin{tabular}{lc|cccc}
  \toprule
  \textbf{Weighting}
  & \textbf{Temporal Scale}
  & \textbf{PSNR$\uparrow$}
  & \textbf{SSIM$\uparrow$}
  & \textbf{LPIPS$\downarrow$}
  & \textbf{ATE RMSE$\downarrow$} \\
  \midrule

  Uniform
  & --
  & 24.81 & 0.81 & 0.19 & 0.284 \\
  \midrule

  Truncated Gaussian
  & Fixed $\sigma=0.20$
  & 24.97 & 0.81 & 0.19 & 0.263 \\

  Truncated Gaussian
  & Fixed $\sigma=0.15$
  & 25.32 & 0.82 & 0.18 & 0.235 \\

  Truncated Gaussian
  & Fixed $\sigma=0.10$
  & 25.75 & 0.82 & 0.18 & 0.203 \\

  Truncated Gaussian
  & Fixed $\sigma=0.05$
  & 26.43 & 0.84 & 0.15 & 0.168 \\

  Truncated Gaussian
  & Fixed $\sigma=0.01$
  & 26.07 & 0.83 & 0.17 & 0.189 \\

  \midrule

  \rowcolor{MyGray}
  \textbf{Truncated Gaussian}
  & \textbf{Time-Adaptive $\boldsymbol{\sigma(\tau)}$}
  &\textbf{27.08}
  &\textbf{0.86}
  &\textbf{0.13}
  &\textbf{0.144} \\
  \bottomrule
  \end{tabular}
  }
\end{table}

\begin{figure}[!t]
  \centering
  \includegraphics[width=0.48\textwidth]{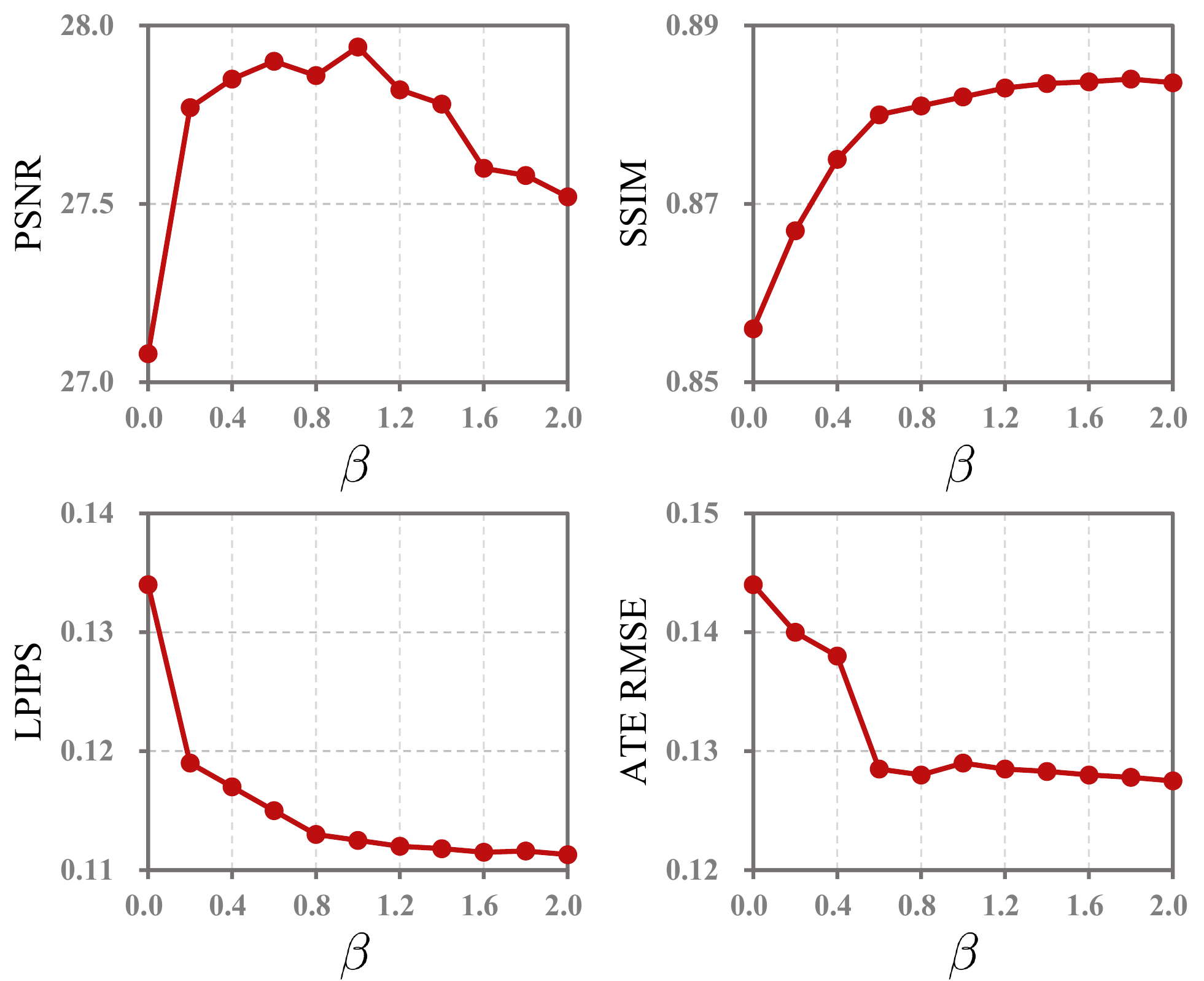}
  \caption{Performance in terms of PSNR, SSIM, LPIPS, and ATE RMSE across different $\beta$ values on the Replica dataset \cite{2019replica}.}
  \label{fig:beta}
\end{figure}

\noindent\textbf{Temporal Weighting and Scale Selection in TCPI.}
We investigate the temporal weighting and scale-selection strategies used by TCPI without using LRES.
As reported in \Cref{tab:tcpi}, uniform weighting and truncated Gaussian weighting with $\sigma=0.20$ achieve similar performance.
Among the fixed-scale variants, progressively decreasing $\sigma$ from $0.20$ to $0.05$ consistently improves both reconstruction quality and pose accuracy.
This indicates that concentrating the integration around the query timestamp reduces the influence of more distant poses while retaining sufficient neighboring information to capture the local motion trend.
However, further reducing $\sigma$ to $0.01$ degrades all metrics.
The overly concentrated distribution weakens the coupling among neighboring trajectory locations, thereby limiting the benefit of temporal integration.
The proposed time-adaptive $\sigma(\tau)$ outperforms all fixed-scale variants, achieving the best performance across all four metrics.
These results suggest that a single fixed integration range cannot adequately accommodate motion variations along the entire trajectory, whereas the predicted temporal scale can adaptively adjust the integration range at different timestamps.

\noindent\textbf{Sensitivity to the Reweighting Factor $\beta$.}
We analyze the sensitivity to $\beta$ in \cref{eq:loss_based_weight}, where $\beta=0$ assigns equal probability to all temporal regions and therefore reduces LRES to uniform sampling.
As shown in \cref{fig:beta}, introducing loss reweighting with $\beta>0$ generally improves both reconstruction quality and pose accuracy over uniform sampling, confirming the benefit of allocating additional samples to intervals with larger reconstruction losses.
As $\beta$ increases, PSNR reaches its maximum around $\beta=1.0$ and decreases slightly for larger values, while SSIM and LPIPS gradually saturate.
ATE RMSE decreases in $\beta\in[0,0.6]$ and remains relatively stable thereafter.
These observations indicate that moderate reweighting sufficiently emphasizes underfitted intervals, whereas an excessively concentrated sampling distribution provides little additional benefit and may reduce temporal coverage.
Overall, $\beta=1.0$ provides the best balance across reconstruction and pose metrics and is used as the default setting.

\section{Conclusion}
In this paper, we propose \textit{EvTrajGS}, an accurate and efficient framework for 3D Gaussian Splatting from unposed event streams.
Specifically, we parameterize camera motion with a continuous-time trajectory function initialized from off-the-shelf pose estimates, providing pose queries at arbitrary timestamps.
Building upon this function, we adaptively aggregate neighboring trajectory states into a temporally coupled pose for rendering, which captures the local motion trends beyond the isolated pose estimates.
Furthermore, we jointly optimize the trajectory and 3DGS representation under event-based supervision, together with a loss-reweighted sampling strategy that adaptively emphasizes underfitted event intervals.
Experiments on synthetic and real-world datasets demonstrate that \textit{EvTrajGS} consistently outperforms state-of-the-art methods in reconstruction quality, pose accuracy, and computational efficiency, breaking the accuracy-efficiency trade-off in unposed event-based 3D reconstruction.

\bibliographystyle{IEEEtran}
\bibliography{egbib}

\vspace{-40pt}
\begin{IEEEbiography}[{\includegraphics[width=1in,height=1.25in,clip,keepaspectratio]{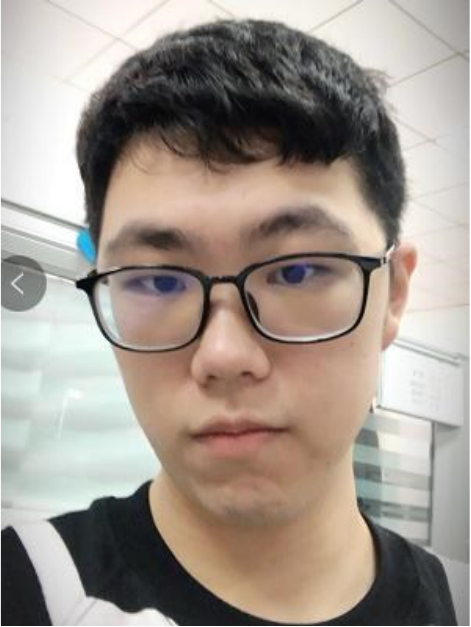}}]{Zixuan Chen}
  received the Ph.D. degree in computer science and engineering from Sun Yat-sen University, China, in 2025.
  He has published over 10 technical papers in international journals and conferences such as IJCV, IEEE TIP, IEEE TCSVT, ICCV and CVPR. 
  His current research focuses on 3D generation and reconstruction.
\end{IEEEbiography}

\vspace{-40pt}
\begin{IEEEbiography}[{\includegraphics[width=1in,height=1.25in,clip,keepaspectratio]{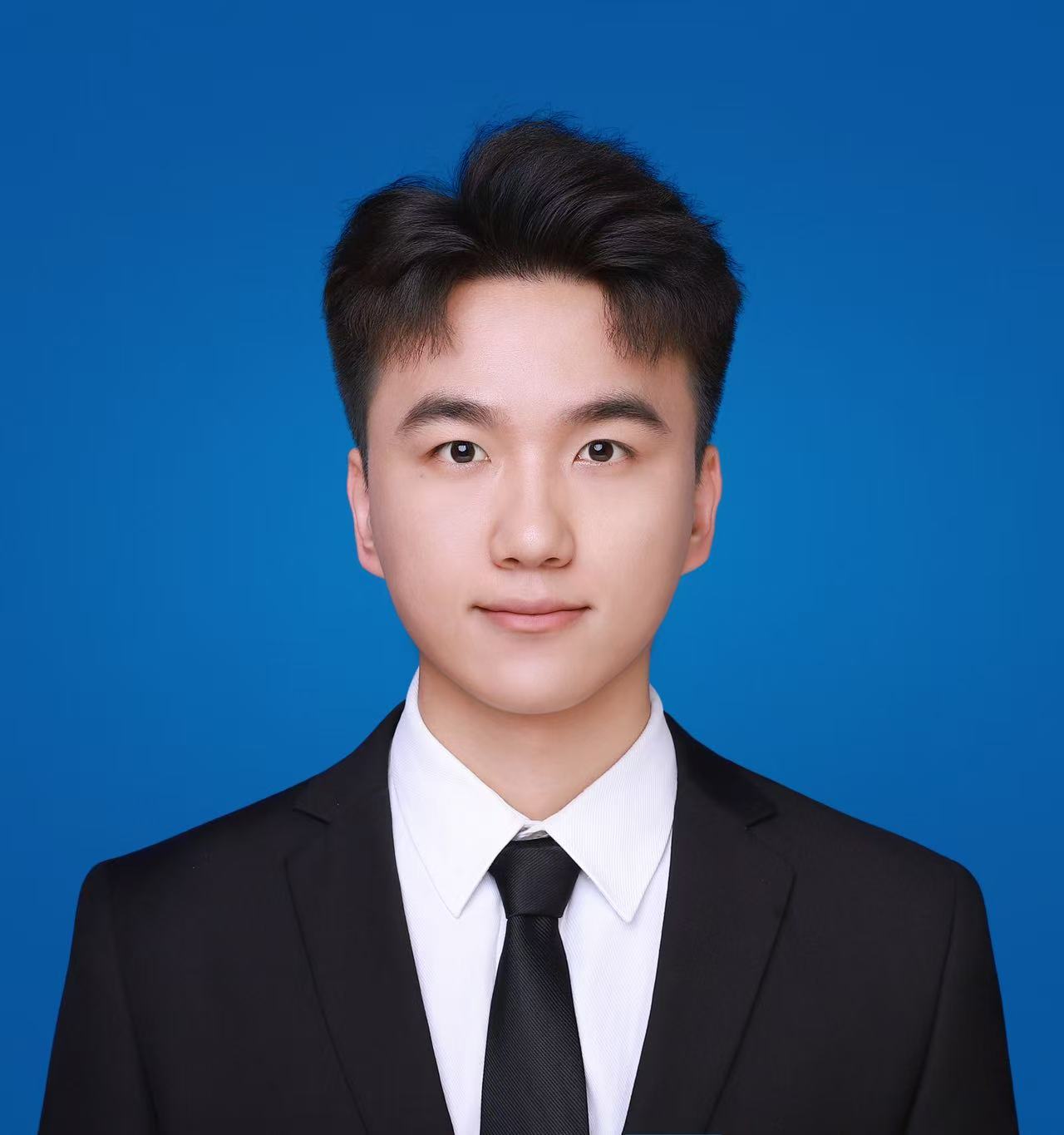}}]{Jiakai Zhang}
 received the B.Eng. degree in Intelligence Science and Technology from Sun Yat-sen University in 2024. 
  He is currently pursuing the M.Eng. degree with the School of Computer Science and Engineering, Sun Yat-sen University. 
  His research interests include 3D reconstruction and vision-language models.
\end{IEEEbiography}

\vspace{-40pt}
\begin{IEEEbiography}[{\includegraphics[width=1in,height=1.25in,clip,keepaspectratio]{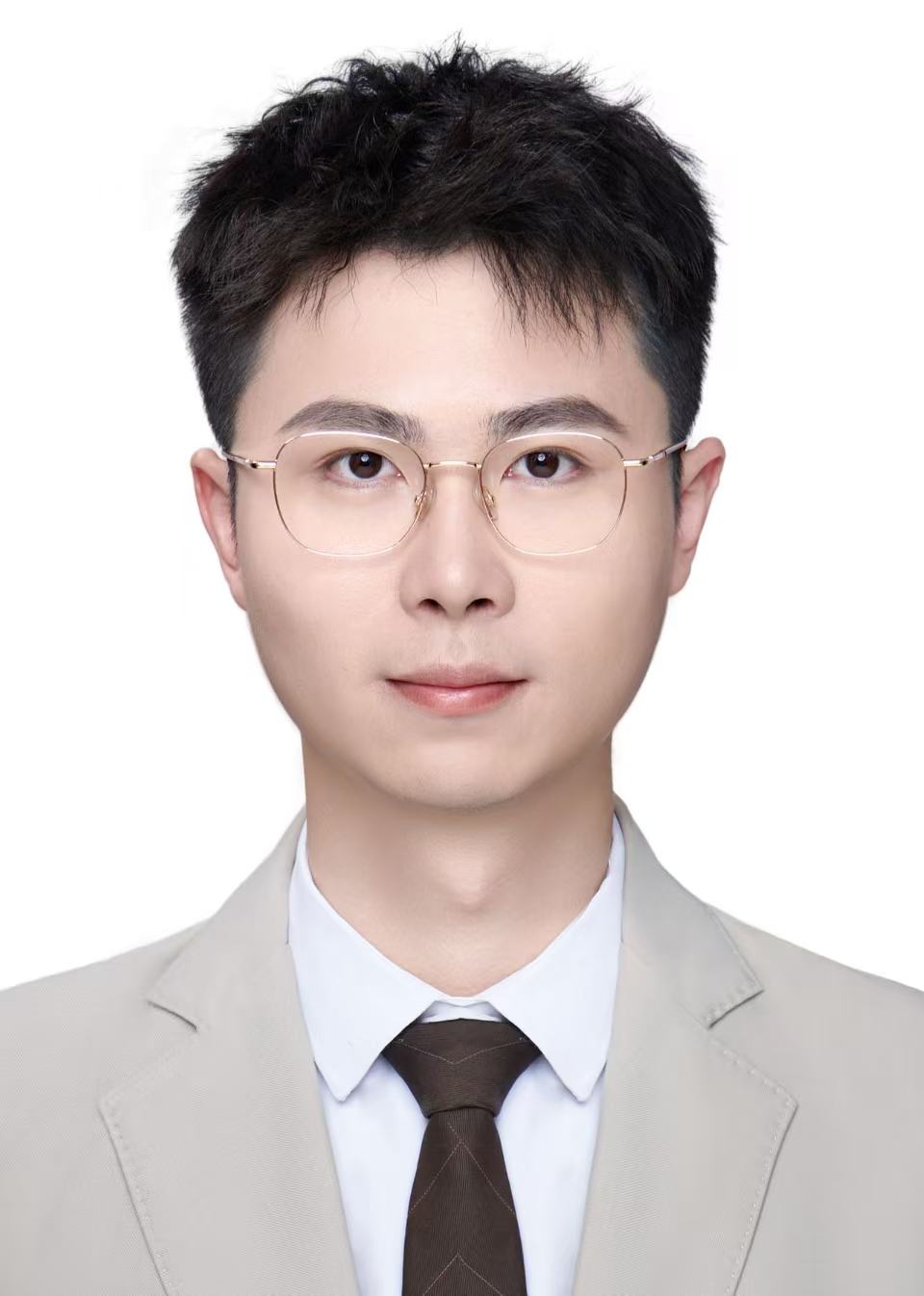}}]{Junhao Dong}
	received the Ph.D. degree from the College of Computing and Data Science, Nanyang Technological University (NTU), Singapore, in 2026, where he is currently a Research Fellow. His research interests include trustworthy AI, computer vision, and machine learning. He received the NTU PhD Medal of Honour and the PREMIA Best Student Paper Award; his paper was selected as a Best Paper Award Candidate at CVPR 2026. He served as an Assistant Program Chair for NeurIPS 2025 and a Session Chair for KDD 2025.
\end{IEEEbiography}

\vspace{-20pt}
\begin{IEEEbiography}[{\includegraphics[width=1in,height=1.25in,clip,keepaspectratio]{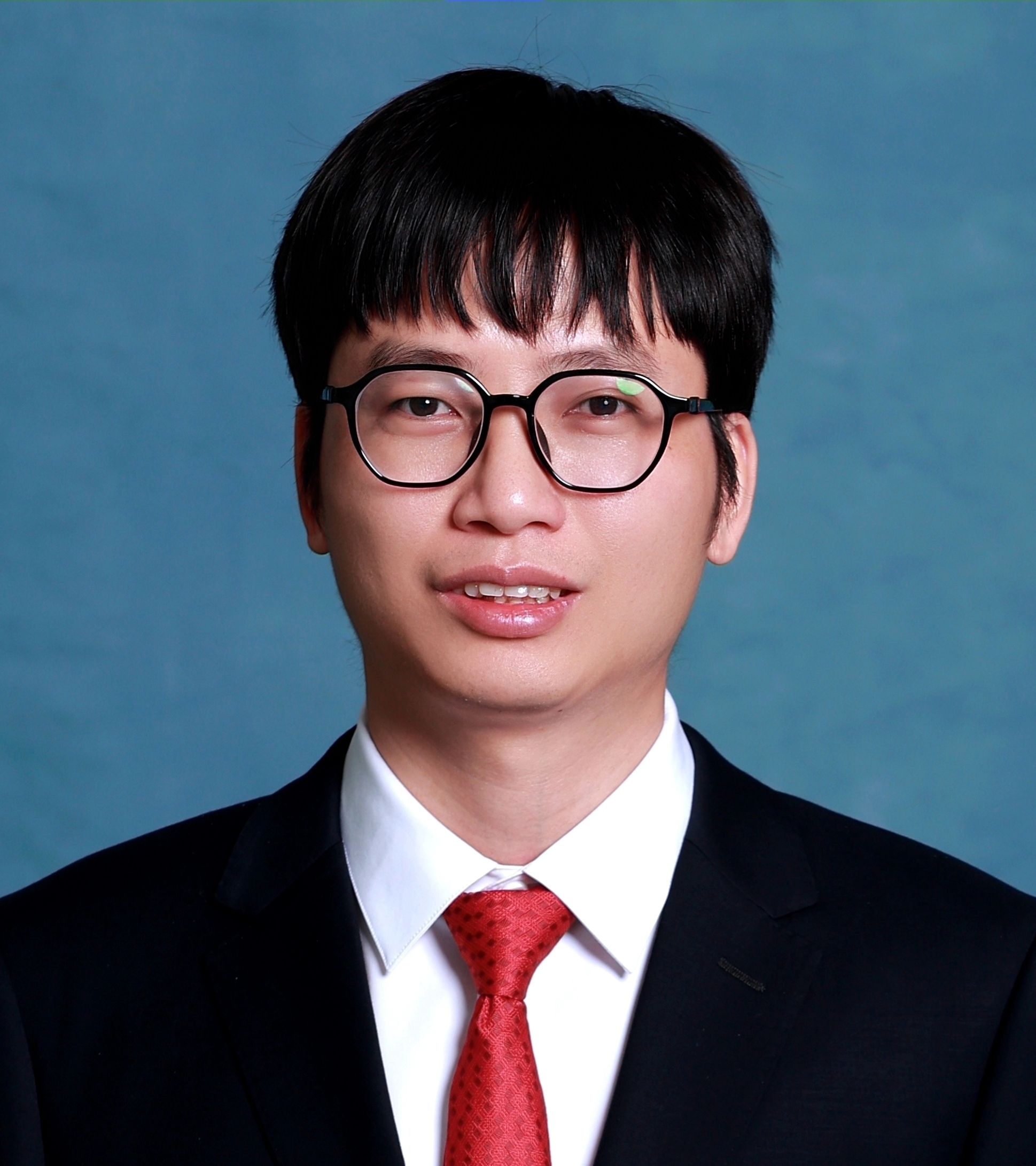}}]{Guangcong Wang}
  received the Ph.D. degree from the School of Computer Science and Engineering, Sun Yat-sen University, Guangzhou, China, in 2020. He is currently an assistant professor with Great Bay University, China. He was a research fellow with the School of Computer Science and Engineering, Nanyang Technological University, Singapore. His research interests include 3D Vision and machine learning. He has published some papers, such as IEEE TPAMI, ACM TOG, ICCV, CVPR, ECCV, NeurIPS, KDD, and AAAI. He served as Area Chair of CVPR, NeurIPS, WACV, BMVC, Senior Program Committee member of AAAI.
\end{IEEEbiography}

\vspace{-20pt}
\begin{IEEEbiography}[{\includegraphics[width=1in,height=1.25in,clip,keepaspectratio]{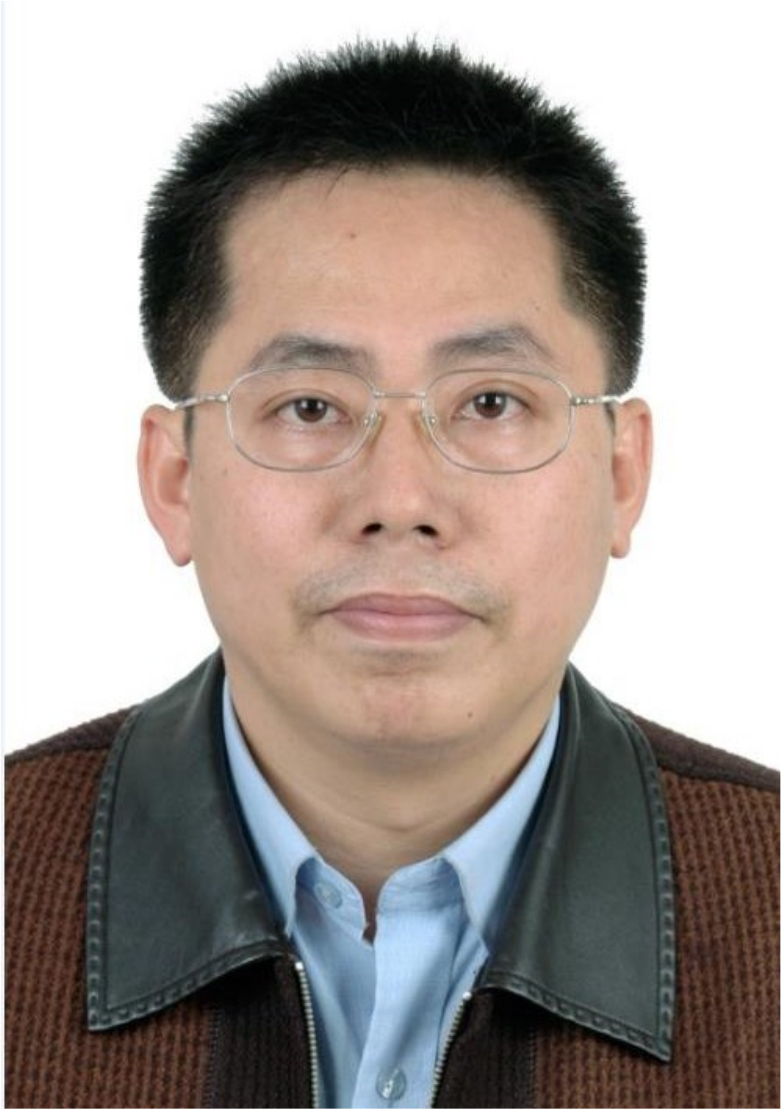}}]{Jianhuang Lai}
  received the Ph.D. degree in mathematics from Sun Yat-sen University, China, in 1999. In 1989, he joined Sun Yat-sen University as an Assistant Professor, where he is currently a Professor of the School of Computer Science and Engineering. He has published over 100 scientific papers in international journals and conferences on image processing and pattern recognition, such as IEEE TPAMI, IEEE TIP, IEEE TNNLS, IEEE TKDE, IEEE TCYB, IEEE TCSVT, PR, ICCV, CVPR, and ICDM. His current research interests are computer vision, digital image processing, pattern recognition, multimedia communication, and multiple-target tracking. He is a fellow of the Image and Graphics Society of China. He serves as the Deputy Director of the Image and Graphics Association of China.
\end{IEEEbiography}

\vspace{-20pt}
\begin{IEEEbiography}[{\includegraphics[width=1in,height=1.25in,clip,keepaspectratio]{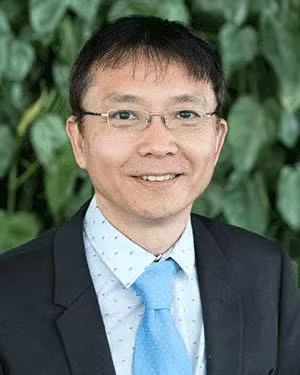}}]{Yew-Soon Ong}
	(Fellow, IEEE) received the Ph.D. degree in artificial intelligence in complex design from the University of Southampton, Southampton, U.K., in 2003. He is the President's Chair Professor of Computer Science with Nanyang Technological University, Singapore, and the Chief Artificial Intelligence Scientist of the Agency for Science, Technology and Research, Singapore. He serves as the Co-Director of Singtel-NTU Cognitive and Artificial Intelligence Joint Lab. His research interest is in artificial and computational intelligence. Dr. Ong is the Founding Editor-in-Chief of the IEEE TETCI and an Associate Editor of IEEE TNNLS, IEEE TCYB, and IEEE TAI. He has received several IEEE outstanding paper awards and was listed as a Thomson Reuters Highly Cited Researcher and among the World's Most Influential Scientific Minds.
\end{IEEEbiography}

\vspace{-20pt}
\begin{IEEEbiography}[{\includegraphics[width=1in,height=1.25in,clip,keepaspectratio]{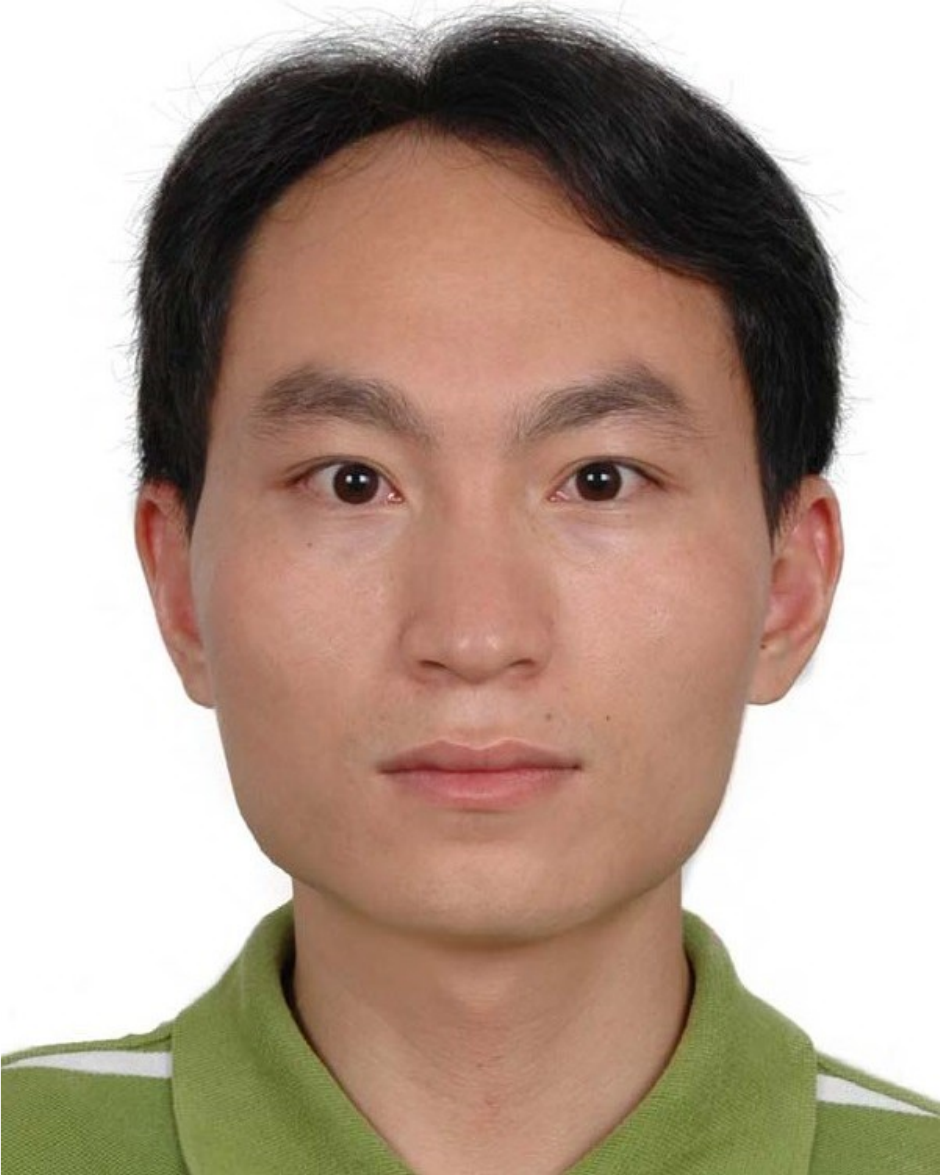}}]{Xiaohua Xie}
  received a Ph.D. degree in applied mathematics from Sun Yat-sen University, China, in 2010. He is currently a Professor at Sun Yat-sen University. He has authored or co-authored over 100 papers in prestigious international journals and conferences. His current research fields cover computer vision, pattern recognition, and AIGC.
\end{IEEEbiography}

\vfill

\end{document}